\documentclass[11pt]{article}
\usepackage[utf8]{inputenc}
\usepackage[margin=1in]{geometry}
\usepackage{amsmath,amssymb}
\usepackage{graphicx}
\usepackage{booktabs,array}
\usepackage[font=small,labelfont=bf,skip=6pt]{caption}
\usepackage[numbers,sort&compress]{natbib}
\usepackage{authblk}
\usepackage{microtype}
\usepackage{xcolor}
\usepackage[hidelinks]{hyperref}
\usepackage{url}
\usepackage[section]{placeins}
\title{Accuracy Is Not Enough: An Efficiency- and Statistics-Aware Benchmark of ResNet50 and Vision Transformer for Explainable Multi-Class Gastrointestinal Diseases Classification}
\author[1,2,3,*]{Akwasi Asare}
\affil[1]{Department of Research and Training, Centre for Medical and Agro Computing (C-MAC), Accra, Ghana}
\affil[2]{Department of Computer Science, Ghana Communication Technology University, Accra, Ghana}
\affil[3]{PerceptronCare Lab.}
\affil[*]{Corresponding author: \href{mailto:nasare34@yahoo.com}{nasare34@yahoo.com}}
\date{}

\begin{document}
\maketitle

\begin{abstract}
Automated classification of gastrointestinal (GI) findings in endoscopic images could support early diagnosis, but architectures are often chosen on accuracy alone. We present a controlled, reproducible benchmark of ResNet50 and ViT-B/16 for eight-class GI classification on the public Kvasir dataset (4,000 images), using a stratified split (400 test images) and stratified group 5-fold cross-validation over all images. In cross-validation, ResNet50 and ViT-B/16 reached 92.12\% and 91.70\% accuracy, and a two one-sided tests (TOST) procedure showed equivalence within a $\pm$3 percentage-point margin (difference +0.43 points, 90\% CI $-$0.33 to +1.18; p \textless{} 0.001). The architectures differed mainly in cost: ResNet50 had 3.6 times fewer parameters and 4.3 times fewer operations, 2.5-fold lower single-image CPU latency (69 vs. 177 ms) and 3.8-fold higher GPU throughput. Cross-validated calibration was similar (expected calibration error 0.021 vs. 0.023). An edge-oriented transformer, MobileViT-S, was most accurate on the test split (95.5\%) but lost 9.4 points when an acquisition artefact present in some classes (the ScopeGuide inset) was occluded, whereas ResNet50 was not significantly affected. On 500 unseen polyp images with expert masks, both models' Grad-CAM maps focused on the lesion (3.6 times chance). On a de-duplicated external subset of GastroVision (1,998 images), accuracy fell to 78--80\% for all three models and differences between them remained small. Accuracy differences between architectures were thus small relative to the effects of shortcut learning and domain shift, whereas efficiency differences were large. An efficient CNN is a defensible default for resource-constrained screening, but no model is deployment-ready without local validation.
\end{abstract}


\section{Introduction}

Gastrointestinal (GI) diseases including inflammatory disorders, ulcers, polyps, and cancers remain among the most prevalent causes of morbidity and mortality worldwide, cause a significant strain on healthcare resources \cite{r1,r2,r3}. The International Agency for Research on Cancer (IARC), the World Health Organization's (WHO) specialized cancer agency, reports that gastrointestinal (GI) cancers are responsible for approximately 3.5 million new cases globally each year. These types of cancers collectively account for an estimated 2.2 million deaths annually, a mortality-to-incidence ratio of about 63\% \cite{r4}. Early and accurate detection of GI abnormalities can dramatically improve treatment outcomes, yet timely diagnosis remains a challenge in many clinical settings due to increasing patient loads, a shortage of trained endoscopists, and the subjective nature of visual interpretation \cite{r5}. Conventional endoscopy, while highly effective for visualizing the internal structure of the gastrointestinal tract, relies heavily on the expertise of clinicians to detect and classify pathological features and it can lead to potential errors in disease classification \cite{r6}. Although endoscopy stands as the undisputed gold standard for inspecting the gastrointestinal (GI) tract \cite{r4}, its true potential is significantly curtailed by a critical variable: operator performance. The efficacy of this vital procedure, therefore, hinges directly on the skill and consistency of the individual wielding the endoscope.  Inter-observer variability and operator fatigue can lead to missed or delayed diagnoses, especially in subtle or early-stage cases. To address these challenges, researchers and clinicians have increasingly turned to artificial intelligence (AI) and deep learning systems to augment diagnostic workflows and reduce human error.

Recent advances in computer vision, particularly convolutional neural networks (CNNs), have demonstrated significant success in medical image classification, segmentation, and anomaly detection tasks \cite{r7,r8,r9}. Architectures such as ResNet, DenseNet, and Inception have been widely adopted in gastrointestinal imaging, showing high performance in tasks like polyp detection\cite{r10} , ulcer identification, and disease staging \cite{r11}. These models excel in learning hierarchical spatial features through convolutional filters, making them well-suited to identifying texture and shape abnormalities in medical images \cite{r5,r11}.

However, despite the success of CNNs, they suffer from certain limitations. CNNs are inherently biased toward local patterns due to their receptive field constraints and inductive biases. This makes them less effective in modeling long-range dependencies or capturing global context an important consideration in medical imaging where subtle variations in texture and anatomical structures may span large regions of the image \cite{r12}. Vision Transformers (ViTs), a recent innovation in the computer vision community, offer a promising alternative. Inspired by transformer models in natural language processing, ViTs utilize self-attention mechanisms to model relationships between all parts of an image simultaneously \cite{r13,r14}. This global reasoning capability enables ViTs to potentially outperform CNNs in tasks requiring holistic interpretation \cite{r12,r15}. However, ViTs generally require larger datasets and are computationally more intensive, raising questions about their practical utility in medical imaging, where labeled data is limited, and real-time processing is often critical \cite{r16}. 

To explore these trade-offs, we conduct a systematic comparison between ResNet50, a widely used, lightweight CNN, and ViT-B/16, a transformer-based architecture, complemented by an edge-oriented hybrid transformer (MobileViT-S), for multi-class gastrointestinal disease classification using the Kvasir dataset. The Kvasir dataset contains 4,000 annotated endoscopic images representing eight clinically relevant classes: \textit{dyed-lifted-polyps} (post-treatment assessment), \textit{dyed-resection-margins} (surgical boundary evaluation), \textit{esophagitis} (inflammation), \textit{normal-cecum} (healthy colon start), \textit{normal-pylorus} (stomach exit), \textit{normal-z-line} (esophagus-stomach junction), \textit{polyps} (pre-cancerous growths), and \textit{ulcerative-colitis} (chronic inflammation). Unlike prior works that focus on binary classification or polyp segmentation, we evaluate these models on a more challenging 8-class classification task that better reflects real-world clinical complexity. In addition to classification accuracy, we assess the models across several critical dimensions: computational efficiency (in terms of MACs and parameter count), generalization (held-out test accuracy, cross-validation and an external dataset), calibration, explainability (Grad-CAM, LIME, SHAP and attention rollout), and robustness to an acquisition artefact. By analyzing both overall metrics and class-specific performance, we aim to uncover the strengths and limitations of each architecture and provide practical guidelines for model deployment in clinical settings.

This paper makes the following key contributions:

\begin{enumerate}

\item We provide a controlled, fully reproducible benchmark of a standard CNN (ResNet50) and a standard Vision Transformer (ViT-B/16) for 8-class gastrointestinal disease classification, evaluated on a fixed stratified split and with stratified group 5-fold cross-validation in which near-duplicate images are never split across folds. Rather than inferring equivalence from a non-significant test, we combine McNemar's paired test with a two one-sided tests (TOST) equivalence procedure and a $\pm$3 percentage-point margin, and show that the two architectures are equivalent in cross-validated accuracy (92.12\% vs. 91.70\%; 90\% CI for the difference $-$0.33 to +1.18 points).

\item We characterize computational cost with a standardized protocol (parameters, multiply-accumulate operations, GPU and CPU latency and throughput, peak memory, and training time). ResNet50 needs 3.6$\times$ fewer parameters and 4.3$\times$ fewer operations, runs 2.5$\times$ faster on a CPU, has 3.8$\times$ higher GPU throughput and reaches its best checkpoint about four times sooner. We complement this with calibration analysis (three ECE binning schemes, class-wise ECE, Brier score and negative log-likelihood, with and without label smoothing), positive predictive values at realistic prevalence, and a multi-method interpretability assessment (Grad-CAM, LIME, SHAP and attention rollout) that includes lesion-level localization against expert polyp masks.

\item We test robustness beyond the curated benchmark. An occlusion experiment shows that an edge-oriented transformer (MobileViT-S), the most accurate model on the test split, depends on an acquisition artefact (the ScopeGuide inset) far more than ResNet50, and an evaluation on a de-duplicated external subset of GastroVision shows a 13--17 point accuracy drop for all models. We translate these findings into deployment-oriented guidance and delineate the limitations that bound them.

\end{enumerate}

The remainder of this paper is organized as follows: Section 2 reviews the related literature; Section 3 details the dataset, model architectures, and training and evaluation methodology; Section 4 presents and discusses the experimental results including classification performance, statistical significance testing, computational efficiency, calibration, interpretability, robustness to an acquisition artefact and external evaluation, together with the limitations of the study; and Section 5 concludes with directions for future research.

\section{Related Work}

The intersection of computer vision and gastrointestinal endoscopy has seen rapid advancements in recent years, driven largely by the growing availability of annotated datasets and improvements in deep learning architectures. Existing literature on GI image analysis can be broadly categorized into three main directions: (1) classical machine learning with handcrafted features, (2) convolutional neural network (CNN)-based models, and (3) emerging transformer-based methods. This section provides a concise overview of each stream, focusing on the context and contributions that inform our study.

\subsection{Early Approaches: Handcrafted Features and Classical Methods}

Initial efforts in automated gastrointestinal diagnosis relied heavily on classical machine learning techniques coupled with manually engineered features. Pogorelov et al. \cite{r17} introduced the Kvasir dataset and evaluated several machine learning classifiers using global features such as texture, color histograms, and Local Binary Patterns (LBP). Among their evaluated methods, the Logistic Model Tree classifier achieved a test accuracy of 93.7\% and an F1-score of 0.747. While promising, these handcrafted approaches lacked generalization ability and struggled with visual variability inherent in endoscopic data. 

\subsection{Deep Learning and CNN-Based Methods}

The advent of deep learning, particularly CNNs, revolutionized medical image analysis by enabling automatic feature learning from raw pixel data. CNNs like AlexNet, VGG, ResNet, and Inception have been widely applied to tasks such as polyp detection, bleeding identification, and disease classification \cite{r7,r8,r18}. Jha et al.\cite{r19} introduced the Kvasir-SEG dataset for polyp segmentation and proposed a ResUNet architecture, achieving a Dice coefficient of 0.788 and mean Intersection over Union (IoU) of 0.778. Although their model performed well in binary segmentation, it was not evaluated on multi-class classification, limiting its clinical scope. Jha et al. \cite{r20} introduced the Kvasir-Instrument dataset for GI tool segmentation and benchmarked it using U-Net, achieving a Dice score of 0.9158 and IoU of 0.8578. While effective overall, the model struggled with specular highlights and multiple instruments, highlighting the need for more robust methods. 

Varalaxmi et al. \cite{r21} utilized ResNet50 CNN for diagnosing gastrointestinal diseases from endoscopy images. Their model, evaluated on the Kvasir dataset, achieved 88.05\% accuracy, 87.05\% sensitivity, and 92.33\% specificity, demonstrating efficient and accurate diagnosis. 

Hannan et al. \cite{r22} achieved 98.9\% accuracy in gastrointestinal disorder diagnosis by combining DenseNet-201 deep feature extraction with a kernel SVM classifier using Kvasir dataset. 

Mary et al. \cite{r23} proposed a CNN-based framework for classifying gastrointestinal diseases using VGG-19 and ResNet-50 on a subset of the Kvasir dataset with 3,500 images across seven classes. ResNet-50 achieved the highest validation accuracy of 96.81\%, while VGG-19 reached 94.21\%. Similarly, Siddiqui et al. \cite{r24} proposed a deep ensemble learning framework for classifying endoscopic gastrointestinal images by combining two state-of-the-art models, NasNet-Mobile and EfficientNet. Trained on the HyperKvasir dataset and evaluated on the Kvasir v1 and Kvasir v2 datasets (containing images across eight classes), the proposed model achieved an accuracy of 97.83\% on Kvasir v1 and 98.45\% on Kvasir v2.

 Demirbaş et al. \cite{r6} proposed a Spatial-Attention ConvMixer (SAC) architecture for classifying gastrointestinal diseases, enhancing the ConvMixer model with a spatial attention mechanism. Evaluated on the Kvasir dataset (containing images across eight classes), the proposed SAC model achieved an accuracy of 93.37\%, outperforming other architectures like Vanilla ViT, Swin Transformer, and the original ConvMixer. Lafraxo and El Ansari \cite{r25} developed ``GastroNet,'' a custom CNN that achieved 96.89\% accuracy on the KVASIR dataset for automated gastrointestinal abnormality detection from endoscopic images.

\subsection{Vision Transformers and Attention-Based Models}

More recently, transformer-based architectures have entered the field of medical imaging, with Vision Transformers (ViTs) showing competitive or superior performance to CNNs on large-scale natural image benchmarks \cite{r15}. Dosovitskiy et al. \cite{r26} proposed the original ViT model, demonstrating that when trained on sufficient data, ViTs could match CNNs in classification accuracy while offering improved attention-based interpretability.

Varam et al. \cite{r14} proposed the on-edge deployment of Vision Transformers (ViTs) for medical diagnostics, applying quantization techniques to optimize model size without compromising performance on the Kvasir-Capsule dataset. Evaluating seven ViT models, including EfficientFormerV2S2 and MobileViT\_V2, they showed that MobileViT\_V2\_175 with F16 quantization achieved the highest F1-score of 0.9534, while MobileViT\_V2\_050 with QAT offered the smallest size (1.70 MB) and best performance-to-size ratio (41.25). Similarly, Regmi et al. \cite{r27}  proposed a study comparing Vision Transformers (ViTs) and Convolutional Neural Networks (CNNs) for efficient medical image classification across three datasets: Chest X-ray, Kvasir, and Kvasir-Capsule. Utilizing various ViT models (e.g., ViT-L/16, ViT-B/16) and data augmentation, they demonstrated that ViT-L/16 achieved the highest F1-scores of 0.9532 for Chest X-ray and 0.9436 for Kvasir. For the Kvasir-Capsule dataset, ViT-B/16 achieved a weighted F1-score of 0.7156. 

Lima et al \cite{r28} proposed using Visual Transformers (ViTs) to classify Video Capsule Endoscopy (VCE) images, aiming to assist in the diagnosis of gastrointestinal tract abnormalities. Evaluating their ViT-based model on the Kvasir-Capsule dataset (11 classes), they reported a high average accuracy of 99.70\%, with a precision of 99.64\%, sensitivity of 99.86\%, and an F1-score of 99.54\%.

Vu et al. \cite{r29} proposed PyBoFormer, a novel deep neural network architecture for polyp segmentation. This model enhances the task by integrating a Transformer encoder (MViT) and a stage-wise feature pyramid decoder structure with a Global-Local Context Aggregation (GLCA) module. Evaluated on multiple polyp datasets, including Kvasir-SEG and ClinicDB, PyBoFormer achieved state-of-the-art performance, with Dice scores of 93.4\% on Kvasir and 96.0\% on ClinicDB. 

Gautam et al. \cite{r30}  introduced Kvasir-VQA-x1, a large-scale multimodal dataset for medical visual question answering (MedVQA) in gastrointestinal (GI) endoscopy, featuring 159,549 complex QA pairs stratified by reasoning difficulty. Designed to test clinical reasoning and robustness, the dataset includes weakly augmented images and supports both standard and perturbed evaluation tracks. Fine-tuned models like Qwen2.5-VL and MedGemma achieved up to 90\% accuracy, but performance declined on higher complexity levels, indicating challenges in multi-hop clinical reasoning. Wang et al. \cite{r31} introduced a Vision Transformer with Hybrid Shifted Windows (HSW Transformer) for gastrointestinal endoscopy image classification. This model addresses the challenge of capturing long-range dependencies in images with widely distributed features, a limitation of prior ViTs like Swin Transformer. Tested on Kvasir v2 and HyperKvasir datasets, HSW Transformer achieved accuracies of 95.42\% and 86.81\% respectively, showcasing its improved ability to capture both short and long-range dependencies. 

Geetha et al. \cite{r32}  proposed a hierarchical Vision Transformer for polyp segmentation, replacing CNN backbones. Their model, with a multi-stage transformer encoder, achieved mean Dice and IoU scores of 74\% and 73\% on the Kvasir-SEG dataset, demonstrating strong performance even on smaller datasets.

\subsection{Beyond White-Light RGB Imaging}

The studies above, like the present work, operate on conventional white-light RGB endoscopic images. A complementary line of research seeks to enrich the information available to the classifier at the imaging stage. Wang et al. \cite{r38} combined hyperspectral imaging with YOLO-based detection to classify superficial esophageal neoplasms (normal esophagus, squamous dysplasia, and squamous cell carcinoma) in white-light and narrow-band images. Huang et al. \cite{r39} proposed the Spectrum-Aided Vision Enhancer (SAVE), a calibrated software transformation that converts white-light images into hyperspectral-like and narrow-band-like representations without dedicated hardware, and reported higher four-class accuracy with SAVE images than with the original images across five CNNs trained on images drawn from Kvasir and the ETIS-Larib polyp database. Chou et al. \cite{r40} applied SAVE to four Kvasir classes (normal mucosa, ulcerative colitis, polyps, and esophagitis) and found that the benefit depended on the architecture: accuracy rose from 85\% to 91\% for VGG16 and from 79\% to 89\% for YOLOv8x, but was unchanged at 94\% for InceptionV3. These approaches seek to expose tissue characteristics that are not visible in RGB and are therefore relevant to the question of clinically meaningful feature extraction examined in Section 4.6. The present study deliberately restricts itself to RGB images, which remain the output of standard endoscopy systems and the modality of public benchmarks such as Kvasir. Whether the accuracy--efficiency conclusions reported here transfer to spectrally enhanced inputs, which add a pre-processing stage and, for hyperspectral inputs, increase input dimensionality and computational cost, remains an open question.

\subsection{Research Gap and Positioning of This Study}

While these transformer- and CNN-based approaches have advanced automated GI disease diagnosis, many focus on narrower tasks such as binary classification or polyp segmentation, and the studies that do compare CNNs and transformers for multi-class GI classification (for instance, Regmi et al. \cite{r27}, Wang et al. \cite{r31}, and Demirbaş et al. \cite{r6}) typically emphasize raw accuracy or propose new architectures rather than jointly characterizing the accuracy--efficiency--interpretability trade-off relevant to deployment. The contribution here is a controlled, deployment-oriented benchmark of an off-the-shelf ResNet50 and ViT-B/16 under an identical, reproducible protocol, in which the comparison is (i) made on a fixed, stratified, seeded test split and replicated with stratified group 5-fold cross-validation; (ii) tested for both difference and equivalence (McNemar's test, the TOST procedure and confidence intervals) rather than reported as bare point estimates; (iii) characterized in terms of computational cost (MACs, parameters, memory, throughput) alongside accuracy, so that the practical cost of any accuracy difference is explicit; and (iv) accompanied by multi-method attribution analysis, a test for reliance on an acquisition artefact, and an external evaluation. The intended scientific contribution is therefore not a new architecture or a new state-of-the-art number, but evidence about whether, once efficiency and statistical uncertainty are considered, a lightweight CNN remains a defensible default for resource-constrained GI screening relative to a standard ViT, a question of direct relevance to low-resource clinical settings.

\section{Methodology}

\begin{figure}[!htbp]
\centering
\includegraphics[width=0.64\linewidth,height=0.8\textheight,keepaspectratio]{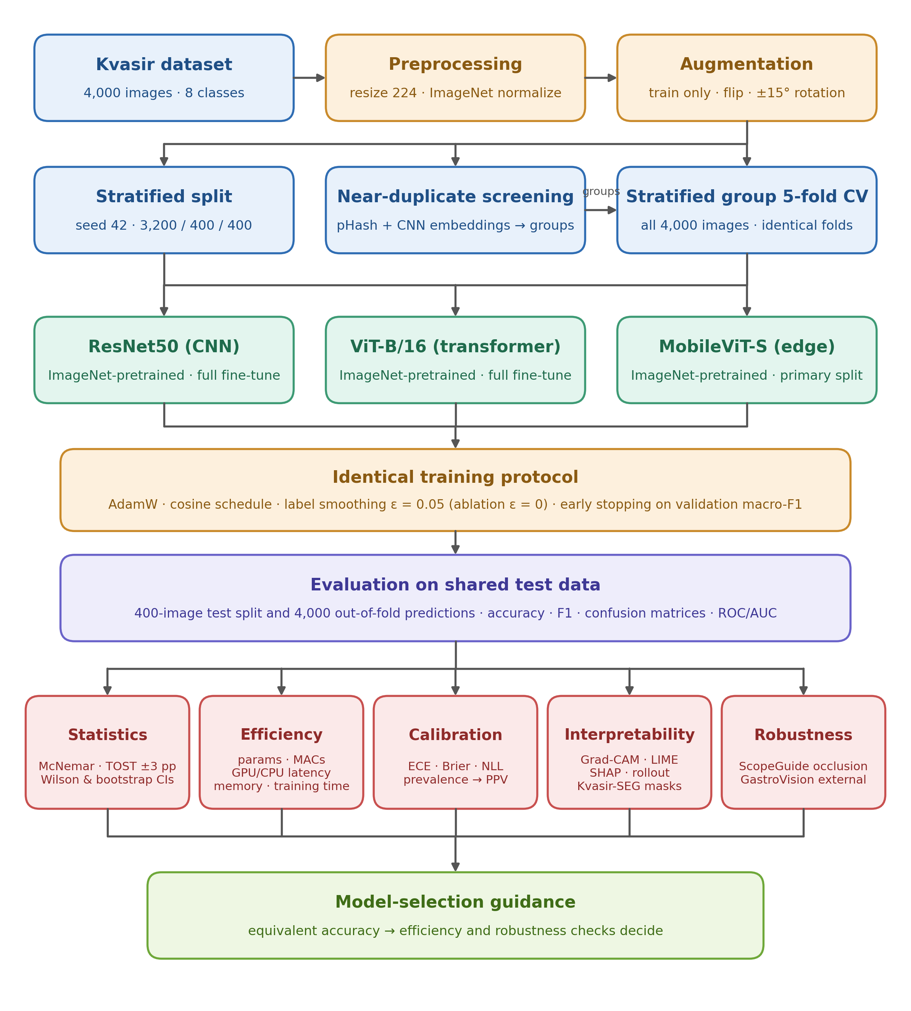}
\caption{Overview of the revised benchmarking pipeline: provenance and near-duplicate screening, preprocessing, the primary stratified split and stratified group 5-fold cross-validation, the three architectures trained under an identical protocol, and the evaluation of accuracy, equivalence, calibration, efficiency, interpretability, shortcut reliance and external generalization.}
\label{fig:1}
\end{figure}

This section describes the dataset, the experimental pipeline, the model architectures compared, the training and evaluation protocol, and the statistical and interpretability methods used. A central design goal of this study is reproducibility: the primary results are derived from a single, fixed, stratified, seeded data split that is shared identically by all models, and are replicated with stratified group 5-fold cross-validation using identical folds for ResNet50 and ViT-B/16, so that any observed difference is attributable to the architectures themselves rather than to differences in data partitioning or training conditions. An overview of the complete pipeline is shown in \textbf{Figure~\ref{fig:1}}.

\subsection{Dataset}

\begin{figure}[!htbp]
\centering
\includegraphics[width=0.58\linewidth,height=0.8\textheight,keepaspectratio]{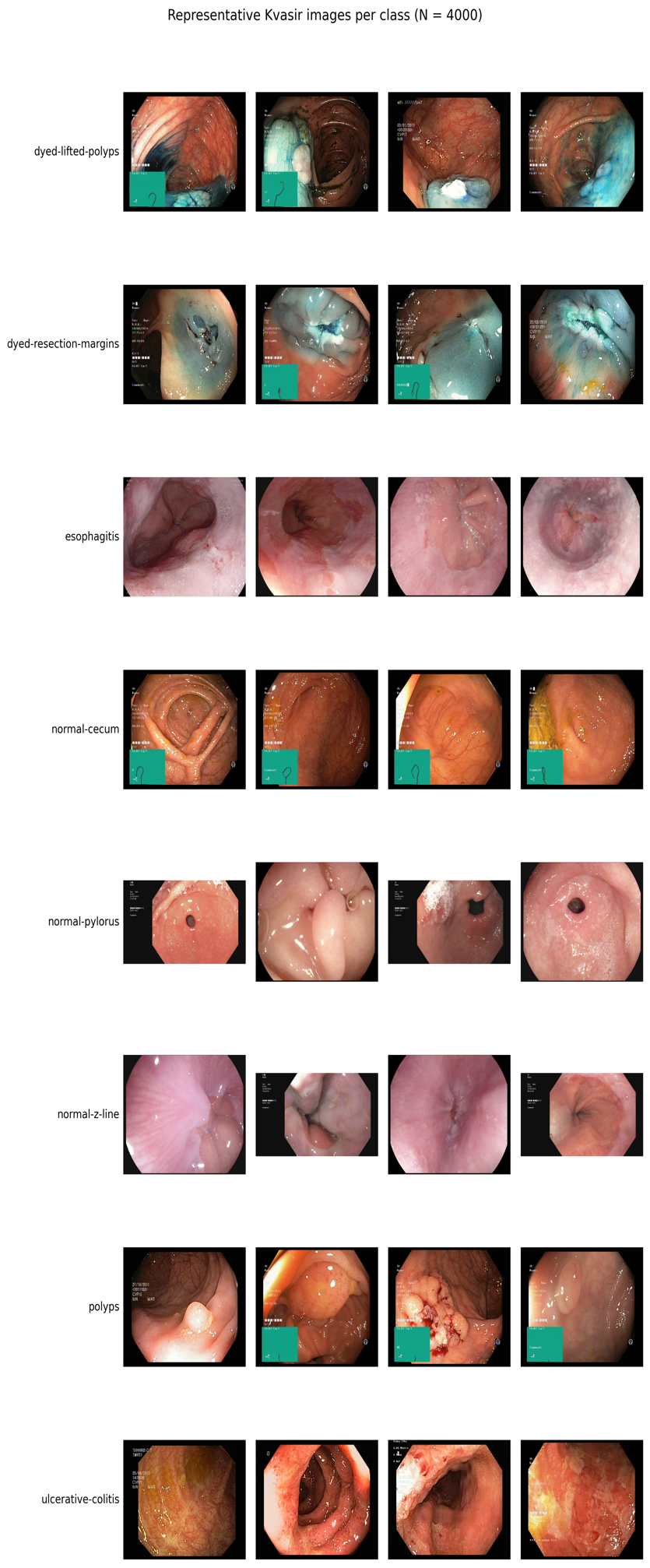}
\caption{Representative endoscopic images (four per class) from each of the eight Kvasir classes.}
\label{fig:2}
\end{figure}

For our experiments, we utilized the Kvasir dataset \cite{r17}, a publicly available collection of verified and annotated endoscopic images of the gastrointestinal tract developed to aid computer-aided diagnosis (CADx) in gastrointestinal endoscopy. We used the original public release (Kvasir v1, obtained from its Kaggle mirror), which comprises 4,000 images with resolutions from 720 $\times$ 576 to 1920 $\times$ 1072 pixels distributed across eight clinically relevant classes: dyed-lifted-polyps, dyed-resection-margins, esophagitis, normal-cecum, normal-pylorus, normal-z-line, polyps, and ulcerative-colitis; which include anatomical landmarks, pathological findings, and procedural scenes: (1) Anatomical landmarks: normal-cecum, normal-pylorus, normal-z-line (2) Pathological findings: esophagitis, polyps, ulcerative-colitis (3) Procedural classes: dyed-lifted-polyps, dyed-resection-margins. Each class contains exactly 500 images (Table~\ref{tab:1}). Representative example images from each class are shown in \textbf{Figure~\ref{fig:2},} illustrating the visual variability across anatomical landmarks and pathological findings. The near-balanced distribution mitigates the risk of the models being biased toward a majority class, although, as discussed in Sections 4.5 and 4.9, it does not reflect the true clinical prevalence of these conditions.

\begin{figure}[!htbp]
\centering
\includegraphics[width=1.0\linewidth,height=0.8\textheight,keepaspectratio]{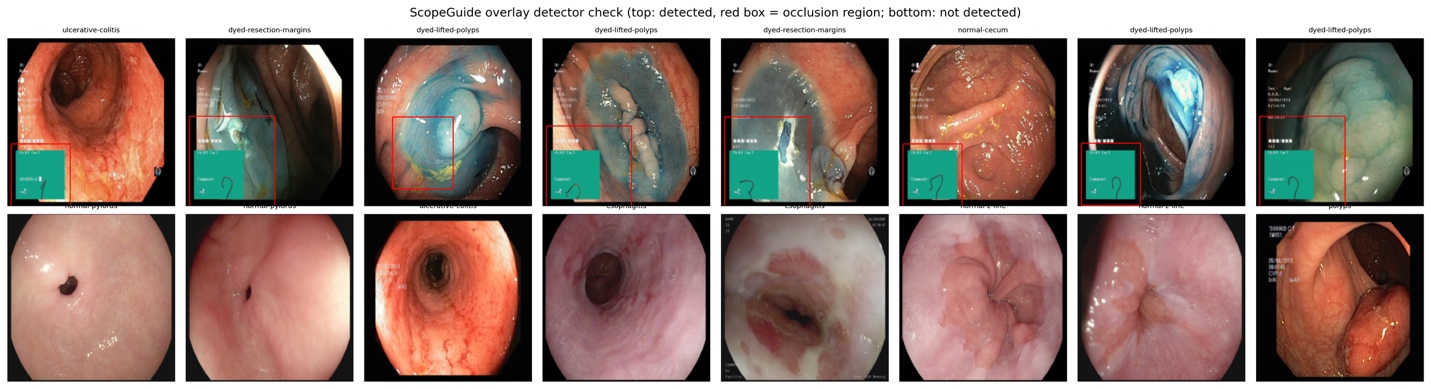}
\caption{The ScopeGuide picture-in-picture inset, an acquisition artefact that occurs only in lower-GI and procedural classes. Top row: images in which the inset was detected automatically (red box: the region blacked out in the shortcut test, Section 4.6); bottom row: images without the inset.}
\label{fig:3}
\end{figure}

\begin{figure}[!htbp]
\centering
\includegraphics[width=0.41\linewidth,height=0.58\textheight,keepaspectratio]{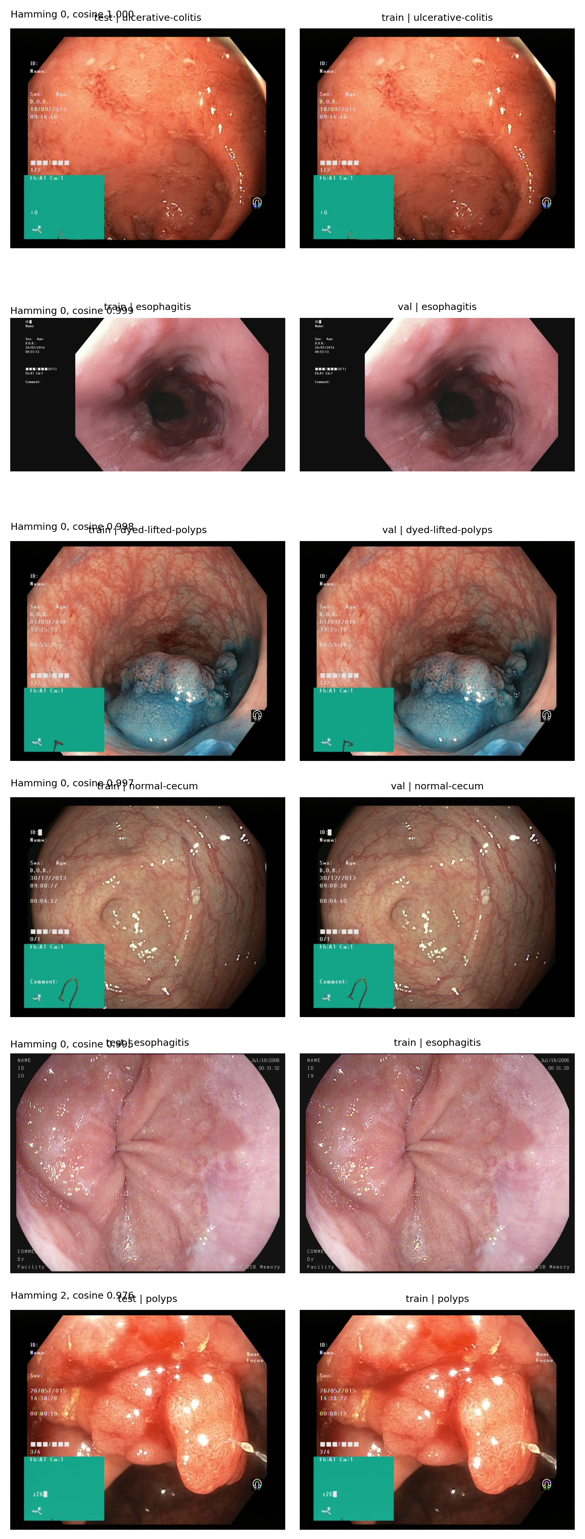}
\caption{The six highest-ranked near-duplicate image pairs that cross the training, validation and test subsets of the primary split, with their perceptual-hash Hamming distance and embedding cosine similarity. These pairs are identical images stored under different file names. All candidate pairs were merged into groups that were kept within a single cross-validation fold, and the 37 affected test images were removed in a sensitivity analysis (Section 4.3).}
\label{fig:4}
\end{figure}

\textbf{Provenance and independence of observations:} Kvasir images were retrieved from routine gastroscopy and colonoscopy examinations at Bærum Hospital (Vestre Viken Hospital Trust, Norway) and labelled by medical experts; this release contains the first 4,000 labelled images and was later extended in Kvasir v2 and HyperKvasir \cite{r4,r17}. The release organises images by class but provides no patient, examination or video identifiers and no EXIF metadata, so a patient- or procedure-level split cannot be constructed. The accessible metadata are summarised in Table~\ref{tab:1}. Upper-GI classes (esophagitis, normal-pylorus, normal-z-line) were mostly acquired at 1280 $\times$ 1024 pixels, whereas lower-GI and procedural classes were mostly 720 $\times$ 576. Many lower-GI images contain a green picture-in-picture inset from the ScopeGuide endoscope-position system (Figure~\ref{fig:3}). Detected automatically, it occurs in 96.6\% of normal-cecum, 71.2\% of dyed-lifted-polyps, 70.2\% of dyed-resection-margins, 50.2\% of polyps and 22.0\% of ulcerative-colitis images, and in none of the upper-GI images. Because the inset is strongly class-dependent but carries no pathological information, it is a candidate shortcut; Section 4.6 tests whether the models rely on it. To limit leakage through visually near-identical images, we screened all image pairs with 64-bit perceptual hashes (Hamming distance $\leq$ 6) and cosine similarity of ImageNet-pretrained ResNet50 embeddings ($\geq$ 0.97), inspected the highest-scoring pairs visually, which confirmed that they are identical images stored under different file names (Figure~\ref{fig:4}), and merged near-duplicates into groups (connected components) that were kept within a single cross-validation fold. Because the combined criterion produced a cluster larger than 1\% of the dataset, only the perceptual-hash criterion was used for grouping, giving 3,771 groups (118 with more than one image; largest 43 images).

\begin{table}[!htbp]
\centering
\small
\setlength{\tabcolsep}{4pt}
\renewcommand{\arraystretch}{1.15}
\caption{Class distribution and accessible provenance metadata of the Kvasir dataset used in this study (no EXIF metadata were present in any image).}
\label{tab:1}
\begin{tabular}{>{\raggedright\arraybackslash}p{0.259\linewidth}>{\raggedright\arraybackslash}p{0.141\linewidth}>{\raggedright\arraybackslash}p{0.184\linewidth}>{\raggedright\arraybackslash}p{0.147\linewidth}>{\raggedright\arraybackslash}p{0.173\linewidth}}
\toprule
\nobreak\hspace{0pt}\textbf{Class} & \nobreak\hspace{0pt}\textbf{Images (\%)} & \nobreak\hspace{0pt}\textbf{Most common resolution} & \nobreak\hspace{0pt}\textbf{Distinct resolutions} & \nobreak\hspace{0pt}\textbf{ScopeGuide inset (\%)} \\
\midrule
\nobreak\hspace{0pt}dyed-lifted-polyps & \nobreak\hspace{0pt}500 (12.5) & \nobreak\hspace{0pt}720 $\times$ 576 & \nobreak\hspace{0pt}2 & \nobreak\hspace{0pt}71.2 \\
\nobreak\hspace{0pt}dyed-resection-margins & \nobreak\hspace{0pt}500 (12.5) & \nobreak\hspace{0pt}720 $\times$ 576 & \nobreak\hspace{0pt}1 & \nobreak\hspace{0pt}70.2 \\
\nobreak\hspace{0pt}esophagitis & \nobreak\hspace{0pt}500 (12.5) & \nobreak\hspace{0pt}1280 $\times$ 1024 & \nobreak\hspace{0pt}3 & \nobreak\hspace{0pt}0.0 \\
\nobreak\hspace{0pt}normal-cecum & \nobreak\hspace{0pt}500 (12.5) & \nobreak\hspace{0pt}720 $\times$ 576 & \nobreak\hspace{0pt}1 & \nobreak\hspace{0pt}96.6 \\
\nobreak\hspace{0pt}normal-pylorus & \nobreak\hspace{0pt}500 (12.5) & \nobreak\hspace{0pt}1280 $\times$ 1024 & \nobreak\hspace{0pt}3 & \nobreak\hspace{0pt}0.0 \\
\nobreak\hspace{0pt}normal-z-line & \nobreak\hspace{0pt}500 (12.5) & \nobreak\hspace{0pt}1280 $\times$ 1024 & \nobreak\hspace{0pt}3 & \nobreak\hspace{0pt}0.0 \\
\nobreak\hspace{0pt}polyps & \nobreak\hspace{0pt}500 (12.5) & \nobreak\hspace{0pt}720 $\times$ 576 & \nobreak\hspace{0pt}3 & \nobreak\hspace{0pt}50.2 \\
\nobreak\hspace{0pt}ulcerative-colitis & \nobreak\hspace{0pt}500 (12.5) & \nobreak\hspace{0pt}720 $\times$ 576 & \nobreak\hspace{0pt}4 & \nobreak\hspace{0pt}22.0 \\
\nobreak\hspace{0pt}Total & \nobreak\hspace{0pt}4,000 (100) & \nobreak\hspace{0pt}--- & \nobreak\hspace{0pt}4 & \nobreak\hspace{0pt}38.8 \\
\bottomrule
\end{tabular}
\end{table}

\subsection{Image Preprocessing and Augmentation and Data Splitting}

All images were resized to 224 x 224 pixels to match the input dimensions expected by both ResNet50 and ViT-B/16, and pixel intensities were normalized using the ImageNet channel means (0.485, 0.456, 0.406) and standard deviations (0.229, 0.224, 0.225) under which the pretrained backbones were trained. To improve generalization without altering the validation or test distributions, data augmentation was applied to the training subset only and consisted of random horizontal flipping and random rotations of up to plus or minus 15 degrees. The validation and test subsets received no augmentation; they used only resizing and normalization. The identical preprocessing and augmentation configuration was applied to both models.

The dataset was partitioned once into training (80\%), validation (10\%), and test (10\%) subsets using stratified sampling, producing 3,200 training, 400 validation and 400 test images (50 test images per class). The split was generated with a fixed random seed (seed = 42) applied to the Python, NumPy and PyTorch random number generators, with cuDNN in deterministic mode, saved to disk, and reused without modification for every model, so that all models were evaluated on exactly the same test images; this is what makes the paired comparisons in Section 4.3 valid.

\textbf{Cross-validation:} To determine whether conclusions depend on this particular partition, we additionally performed stratified group 5-fold cross-validation over all 4,000 images, with the near-duplicate groups of Section 3.1 never split across folds. In each fold the held-out fold (777--842 images) served as the test set and the remaining images were divided 8:1 into training and validation subsets, again stratified and group-aware. ResNet50 and ViT-B/16 were trained on identical folds with the protocol of Section 3.3 and a fold-specific seed (42 + fold index). Pooling the out-of-fold predictions yields one paired prediction per image for each model (n = 4,000), the basis of the confirmatory analyses in Section 4.3.

\subsection{Model Architectures and Training Configuration}

Convolutional neural networks (CNNs) were among the earliest deep learning architectures to achieve clinically meaningful performance in medical image classification and segmentation \cite{r33,r34,r35}. We compare two representative, widely used architectures, each initialized with ImageNet-pretrained weights (torchvision IMAGENET1K\_V2 for ResNet50 and IMAGENET1K\_V1 for ViT-B/16) so that both benefit from large-scale natural-image features before fine-tuning on the gastrointestinal data. The two backbones embody fundamentally different inductive biases local convolution versus global self-attention which is precisely the contrast this study sets out to evaluate.

\paragraph{ResNet50 (convolutional neural network).}

\begin{figure}[!htbp]
\centering
\includegraphics[width=0.97\linewidth,height=0.8\textheight,keepaspectratio]{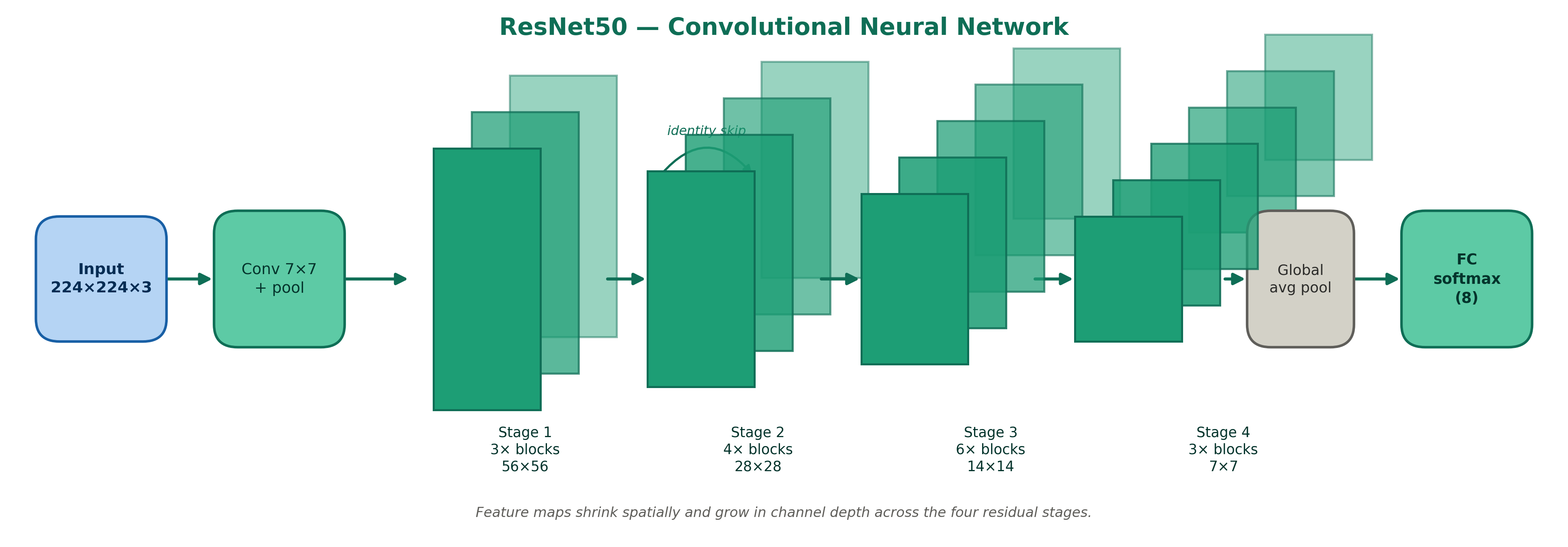}
\caption{ResNet50 architecture --- convolutional stem, four stages of bottleneck residual blocks with identity skip connections, global average pooling, and an 8-way classification head.}
\label{fig:5}
\end{figure}

ResNet50 \cite{r36,r37} is a 50-layer convolutional neural network organized into a stem (a 7$\times$7 convolution followed by max pooling) and four sequential stages of bottleneck residual blocks, with progressively increasing channel depth and decreasing spatial resolution. Each residual block applies a sequence of 1$\times$1, 3$\times$3, and 1$\times$1 convolutions and adds the block's input back to its output through an identity skip connection; these skip connections mitigate the vanishing-gradient problem and allow stable optimization of very deep networks. Because convolutions operate over local receptive fields with shared weights, the network learns hierarchical, locally biased spatial features --- edges and textures in early stages, and larger structures and object-level patterns in deeper stages while remaining comparatively lightweight (23.5M parameters). After the final stage, a global average-pooling layer collapses the feature maps into a single feature vector that is passed to the classifier. For our 8-class task, the original 1000-way ImageNet head was replaced with an 8-way fully connected layer, and all layers were fine-tuned end-to-end. The overall architecture is illustrated in Figure~\ref{fig:5}.

\paragraph{ViT-B/16 (Vision Transformer).}

\begin{figure}[!htbp]
\centering
\includegraphics[width=0.97\linewidth,height=0.8\textheight,keepaspectratio]{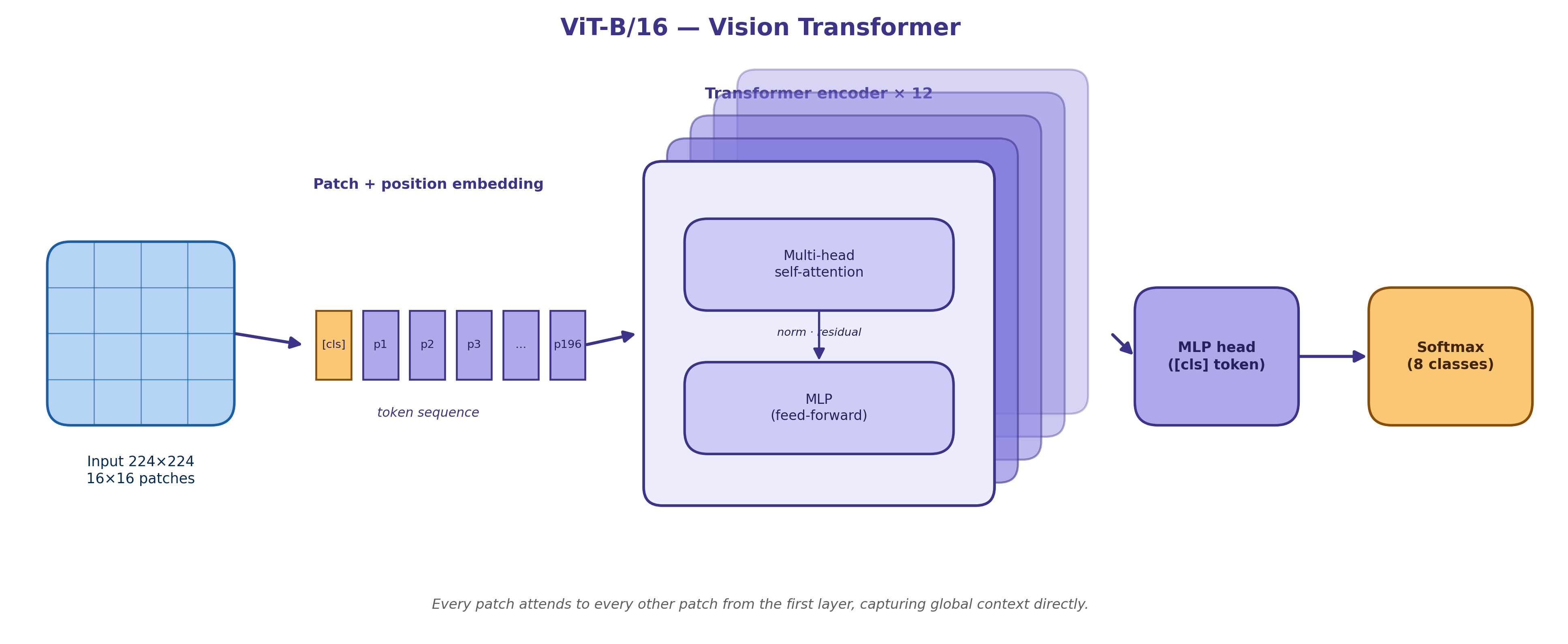}
\caption{ViT-B/16 architecture --- patch embedding of 16$\times$16 patches with positional encoding, twelve transformer encoder blocks (multi-head self-attention and MLP), and a classification-token MLP head producing the 8-way output.}
\label{fig:6}
\end{figure}

ViT-B/16 \cite{r26} is the base Vision Transformer operating on 16$\times$16 image patches. The 224$\times$224 input is divided into a grid of non-overlapping 16$\times$16 patches (196 patches in total); each patch is flattened and linearly projected into a token embedding, a learnable position embedding is added to retain spatial order, and a special classification token is prepended to the sequence. The resulting token sequence is processed by a stack of twelve transformer encoder blocks, each consisting of multi-head self-attention followed by a feed-forward multilayer perceptron, with layer normalization and residual connections around both sub-layers. Unlike convolution, self-attention lets every patch attend to every other patch from the very first layer, so the model can capture global, long-range dependencies directly rather than building them up gradually through depth; this expressiveness comes at a higher computational and parameter cost (85.8M parameters). Classification is performed from the final-layer classification token through an MLP head. We used the torchvision ViT-B/16 model, replaced its head with an 8-way linear layer, and fine-tuned the network end-to-end under the same protocol as ResNet50. The overall architecture is illustrated in Figure~\ref{fig:6}.

Both models were trained under an identical optimization protocol to ensure a fair comparison. Training used the AdamW optimizer with an initial learning rate of 1 x 10\textsuperscript{-4} and weight decay of 1 x 10\textsuperscript{-4}, a cosine-annealing learning-rate schedule, a batch size of 32, and cross-entropy loss with label smoothing of 0.05. Models were trained for up to 30 epochs with early stopping (patience of 7 epochs) monitoring the validation macro-averaged F1-score; the checkpoint achieving the best validation macro-F1 was retained for all subsequent evaluation. Under this policy ResNet50 reached its best validation macro-F1 (0.935) at epoch 8 and stopped at epoch 15 (39 s per epoch, 10.1 min in total), whereas ViT-B/16 reached its best value (0.935) at epoch 20 and stopped at epoch 27 (118 s per epoch, 55.2 min). All experiments were run in PyTorch 2.10 on a single NVIDIA Tesla T4 GPU (16 GB) in a Kaggle notebook; images were decoded and resized once and cached in memory, and data loading used no worker processes.

\textbf{Label-smoothing ablation:} Because label smoothing affects calibration \cite{r41}, both models were also retrained on the primary split with label smoothing disabled ($\varepsilon$ = 0), all other settings unchanged (Section 4.5). \textbf{Edge-optimized transformer:} To examine whether a transformer designed for resource-constrained devices changes the picture, we also fine-tuned MobileViT-S \cite{r42} (timm implementation, ImageNet-pretrained, 4.9M parameters) on the primary split under the same protocol. ViT-B/16 was retained as the main transformer baseline because it is the canonical and most widely reported vision transformer, including in prior Kvasir benchmarks \cite{r27}, so the main comparison answers the question practitioners most often face: whether a standard pretrained transformer should replace a standard pretrained CNN.

\subsection{Evaluation Metrics}

Model performance was assessed exclusively on the held-out test set of 400 images, which contains no image seen during training or validation, so that all reported figures reflect generalization rather than memorization. We report overall accuracy, and per-class precision, recall, and F1-score, together with macro-averaged values. Precision is the fraction of predicted positives that are correct; recall (sensitivity) is the fraction of true positives that are correctly identified; and the F1-score is their harmonic mean. We additionally report multi-class ROC curves with per-class and micro-averaged area under the curve (AUC), and we summarize discriminative confusion structure using confusion matrices. To quantify uncertainty, 95\% confidence intervals are reported for overall accuracy (Wilson score interval) and for per-class F1 (non-parametric bootstrap with 10,000 resamples).

Formally, for each class these metrics are defined in terms of the counts of true positives (TP), true negatives (TN), false positives (FP), and false negatives (FN) as given in Equations~\eqref{eq:1}--\eqref{eq:5}. Accuracy is the proportion of all predictions that are correct; precision is the proportion of positive predictions that are correct; recall (sensitivity) is the proportion of actual positives that are recovered; specificity is the proportion of actual negatives correctly identified; and the F1-score is the harmonic mean of precision and recall. For the multi-class setting, the macro-averaged F1 is the unweighted mean of the per-class F1-scores over the $C = 8$ classes (Equation~\eqref{eq:6}), which weights every class equally regardless of its support. Model calibration is quantified by the expected calibration error (ECE; Equation~\eqref{eq:7}), which bins the $n$ test predictions into $M = 15$ equal-width confidence bins $B_{m}$   and accumulates the weighted gap between the accuracy and the mean confidence within each bin \cite{r43}. Because ECE depends on the binning scheme, we also report ECE with 10 equal-width and 15 equal-mass bins, class-wise ECE \cite{r44}, the multi-class Brier score \cite{r45} and the negative log-likelihood (NLL), defined in Equations~\eqref{eq:8} and \eqref{eq:9}, each with a 95\% bootstrap confidence interval (10,000 resamples), and paired bootstrap intervals for the difference between models

\begin{equation}
\text{Accuracy} = \frac{TP +  TN}{TP +  TN +  FP +  FN}
\label{eq:1}
\end{equation}

\begin{equation}
\text{Precision} = \frac{TP}{TP + FP}
\label{eq:2}
\end{equation}

\begin{equation}
\text{Recall}\, = \frac{TP}{TP +  FN}
\label{eq:3}
\end{equation}

\begin{equation}
\text{Specificity} = \frac{TN}{TN +  FP}
\label{eq:4}
\end{equation}

\begin{equation}
F_{1} = \frac{2 \times \text{Precision} \times \text{Recall}}{\text{Precision} + \text{Recall}}
\label{eq:5}
\end{equation}

\begin{equation}
F_{1,\text{macro}} = \frac{1}{C}\sum_{i = 1}^{C}F_{1,i}
\label{eq:6}
\end{equation}

\begin{equation}
\text{ECE} = \sum_{m = 1}^{M}{\frac{\left| B_{m} \right|}{n}\left| \text{acc}\left( B_{m} \right) - \text{conf}\left( B_{m} \right) \right|}
\label{eq:7}
\end{equation}

where $C$ is the number of classes ($C = 8$); $F_{1,i}$ is the F1-score of class $i$; $M$ is the number of confidence bins; $B_{m}$ is the set of predictions in bin $m$; $n$ is the total number of test samples; and $\mathrm{acc}(B_{m})$ and $\mathrm{conf}(B_{m})$ are, respectively, the mean accuracy and mean predicted confidence of the predictions in bin $m$.

\begin{equation}
\text{Brier} = \frac{1}{n}\sum_{i = 1}^{n}{\sum_{k = 1}^{C}\left( p_{ik} - y_{ik} \right)^{2}}
\label{eq:8}
\end{equation}
\begin{equation}
\text{NLL} = - \frac{1}{n}\sum_{i = 1}^{n}{\log p_{i,y_{i}}}
\label{eq:9}
\end{equation}

where $p_{ik}$ is the predicted probability of class $k$ for image $i$, $y_{ik}$ equals 1 if image $i$ belongs to class $k$ and 0 otherwise, and $p_{i,y_i}$ is the probability assigned to the true class of image $i$.

\subsection{Statistical Comparison}

Because the two models were evaluated on the same fixed test set, their predictions are paired and can be compared with McNemar's test, the standard paired test for two classifiers on a shared sample. McNemar's test is computed on the 2 x 2 contingency table of cases where the two models agree or disagree, using the discordant pairs (cases correct for one model but not the other). We report both the continuity-corrected chi-square statistic and the exact binomial p-value, along with a 95\% confidence interval for the difference in accuracy obtained by paired bootstrap resampling. This allows us to test whether any observed accuracy difference is statistically meaningful rather than attributable to sampling variation.

Absence of a significant difference does not demonstrate equivalence. We therefore tested equivalence explicitly with the two one-sided tests (TOST) procedure \cite{r46} applied to the paired accuracy difference $\Delta = \mathrm{Acc}(\text{ResNet50}) - \mathrm{Acc}(\text{ViT})$. Equivalence is declared at $\alpha = 0.05$ when the 90\% confidence interval for $\Delta$ lies entirely within $(-\delta, +\delta)$. We set $\delta = 3$ percentage points, i.e. we regard accuracy differences smaller than three images per hundred as not clinically meaningful for an eight-class decision-support tool whose outputs are reviewed by an endoscopist. The margin was fixed before the cross-validation was run, which is therefore the confirmatory analysis. So that readers can apply a different margin, we also report the smallest margin at which equivalence would be declared (the larger absolute bound of the 90\% CI). TOST p-values use the paired standard error of Equation~\eqref{eq:10}, where $b$ and $c$ are the numbers of images classified correctly by only one of the two models and $n$ is the number of test images.

\begin{equation}
\text{SE}(\Delta) = \frac{\sqrt{b + c - \frac{(b - c)^{2}}{n}}}{n}
\label{eq:10}
\end{equation}

\subsection{Computational Efficiency Profiling}

To characterize deployment cost, we profiled each model's number of trainable parameters, multiply-accumulate operations (MACs) for a single 224 x 224 input, and inference latency, throughput and peak memory on GPU and CPU under the standardized protocol described below. These metrics quantify the practical price of each architecture independently of accuracy and are central to the model-selection question this study addresses.

Every measurement ran in a fresh process with the model in evaluation mode, gradients disabled (torch.inference\_mode), random input tensors, and image decoding and pre-processing excluded. On the GPU (NVIDIA Tesla T4, 16 GB, CUDA 12.8) we ran 50 warm-up iterations followed by 200 timed iterations, repeated 5 times, in FP32 and FP16 (autocast), with torch.cuda.synchronize() around every timed call. On the CPU (Intel Xeon 2.0 GHz, 4 threads, 33.7 GB RAM) we used 10 warm-up and 50 timed iterations $\times$ 3 repeats at batch size 1, and 2 warm-up and 5 timed iterations $\times$ 3 repeats at batch size 32. We report single-image latency (batch size 1; median and interquartile range), throughput at batch size 32, peak GPU memory (torch.cuda.max\_memory\_allocated) and CPU inference memory, measured as the peak resident memory sampled every 2 ms during inference minus the resident memory immediately before it. Parameters and multiply-accumulate operations were counted with ptflops. Training cost (seconds per epoch and time to the best checkpoint) was taken from the cross-validation runs.

\subsection{Interpretability Analysis}

We examined model decision-making using Gradient-weighted Class Activation Mapping (Grad-CAM), which produces class-discriminative heatmaps highlighting the image regions most influential to a prediction. For ResNet50, Grad-CAM was computed at the final convolutional block; for ViT, it was computed at the final transformer block with an appropriate token-to-grid reshape. The maps are interpreted qualitatively; the energy-in-field-of-view score used in the original submission was dropped, because it only measures whether attribution avoids dark borders and overlays, and was replaced by the lesion-level analysis below.

To triangulate these findings, we additionally generated LIME (superpixel-based) and SHAP (gradient-based) attributions for both models and attention-rollout maps for ViT. Consistent with Section 4.9, these maps are treated as exploratory evidence of clinical plausibility rather than as proof of clinically valid reasoning.

\textbf{Lesion-level localization:} To test whether attribution falls on disease-associated mucosa rather than merely inside the field of view, we used Kvasir-SEG \cite{r19}, which provides pixel masks, drawn by a medical doctor and verified by an experienced gastroenterologist, for 1,000 polyp images from Kvasir v2. Few of these images are in the Kvasir v1 release used here (9 near-identical by perceptual hash), so we evaluated localization on Kvasir-SEG images not seen in training: the 107 images within Hamming distance 10 of any training or validation image were excluded, leaving 893 eligible images, of which a random sample of 500 was analysed. The class polyps was explained with Grad-CAM (both models) and ViT attention rollout on all 500 images, and with SHAP GradientExplainer and LIME (both models) on 50 of them, because these two methods are computationally expensive. We computed the pointing-game hit rate \cite{r47} (whether the attribution maximum lies inside the lesion), the energy-in-lesion fraction (Equation~\eqref{eq:11}, where $A$ is the non-negative attribution map, $M$ the binary lesion mask and $\odot$ element-wise multiplication; its chance level equals the lesion's area fraction), and the IoU of the top-20\% attribution region with the mask. \textbf{Shortcut test.} To test whether models rely on the ScopeGuide inset (Section 3.1), we blacked out the inset in every test image containing it and compared each model's predictions before and after occlusion with McNemar's test. For 40 of these images we also measured the share of attribution mass that each method placed on the inset, compared with the inset's area fraction.

\begin{equation}
E_{\text{lesion}} = \frac{\sum (A \odot M)}{\sum A}
\label{eq:11}
\end{equation}

\subsection{External Evaluation and Prevalence Analysis}

\textbf{External evaluation:} GastroVision \cite{r48} contains 8,000 endoscopic images in 27 classes acquired at Bærum Hospital (Norway) and Karolinska University Hospital (Sweden). Seven of its classes correspond to Kvasir classes (cecum, pylorus, gastroesophageal junction/normal Z-line, esophagitis, colon polyps, dyed-lifted polyps and dyed-resection margins; it has no ulcerative-colitis class), giving 2,150 images. Because part of GastroVision comes from the same hospital as Kvasir, we removed the 152 images within Hamming distance 6 of any Kvasir image, leaving 1,998 images. The primary-split models were applied unchanged. GastroVision is therefore a partially independent, naturally imbalanced external set rather than a fully independent validation.

\textbf{Prevalence: }To examine behaviour at clinical prevalence, we computed positive predictive values (PPV) at prevalences of 1--20\% from each model's cross-validated sensitivity and false-positive rate, and compared calibration on the out-of-fold predictions re-weighted to an illustrative screening-like class prior (75\% normal landmarks), with and without prior-shift correction of the softmax outputs.

\section{Results and Discussion}

This section reports the outcomes on the shared, stratified test set of 400 images and in cross-validation. We present classification performance (Section 4.1), confusion and ROC analysis (4.2), the statistical comparison including the cross-validation (4.3), computational efficiency, training cost and convergence (4.4), calibration and predictive values under prevalence shift (4.5), interpretability and the shortcut test (4.6), and the external evaluation (4.7), followed by the discussion (4.8) and limitations (4.9).

\subsection{Classification Performance}

\begin{figure}[!htbp]
\centering
\includegraphics[width=1.0\linewidth,height=0.4\textheight,keepaspectratio]{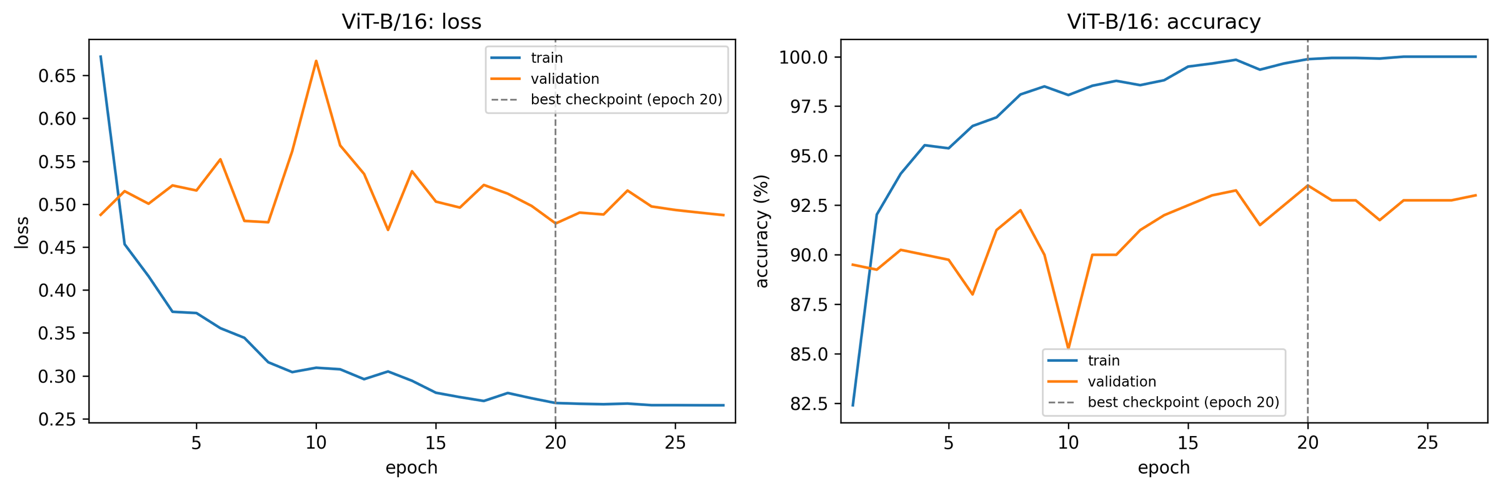}
\\[4pt]
\includegraphics[width=1.0\linewidth,height=0.4\textheight,keepaspectratio]{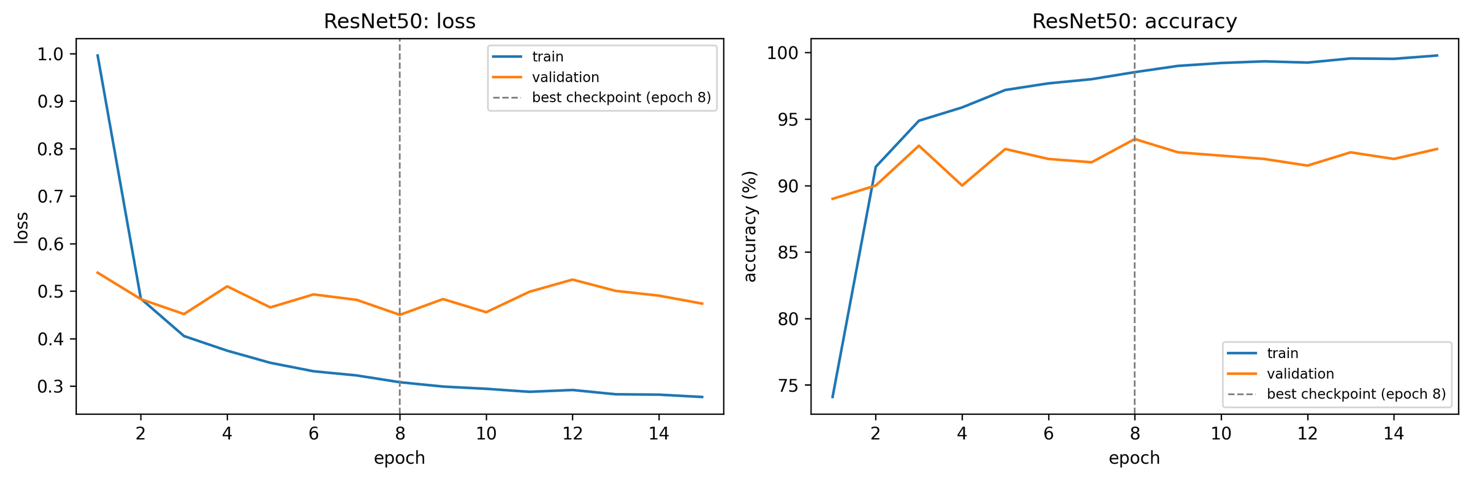}
\caption{Training and validation loss and accuracy curves of ViT-B/16 (upper panels; best validation macro-F1 0.935 at epoch 20, early stopping at epoch 27) and ResNet50 (lower panels; best validation macro-F1 0.935 at epoch 8, early stopping at epoch 15).}
\label{fig:7}
\end{figure}

On the held-out test set, ResNet50 achieved an overall accuracy of 93.75\% (375 of 400 images) and ViT-B/16 92.25\% (369 of 400), a difference of six images; macro-averaged F1 was 0.937 and 0.922. MobileViT-S reached 95.50\% (382 of 400; macro-F1 0.955). The training and validation curves of the two main models are shown in Figure~\ref{fig:7}. ResNet50 reached its best validation macro-F1 at epoch 8, whereas ViT required 20 epochs and its validation curve fluctuated more (for example, validation accuracy dropped to 85.3\% at epoch 10). Both models approached 100\% training accuracy, with a train--validation gap of about 5--7 percentage points that the early-stopping and best-checkpoint policy controls. The per-class precision, recall, and F1-scores are reported in Table~\ref{tab:2} (ViT) and Table~\ref{tab:3} (ResNet50).

As shown in Table~\ref{tab:2}, ViT attained perfect recall for normal-cecum and normal-pylorus; its lowest recall and F1 were for esophagitis (recall 84.0\%, F1 0.848) and normal-z-line (F1 0.851). As shown in Table~\ref{tab:3}, ResNet50 attained perfect recall for dyed-resection-margins, normal-cecum and normal-pylorus and perfect precision for dyed-lifted-polyps and ulcerative-colitis; its lowest recall was for esophagitis (84.0\%) and dyed-lifted-polyps (86.0\%). For both models the weakest classes were esophagitis and normal-z-line (F1 0.85--0.86), whereas the normal landmarks exceeded 0.97.

\begin{table}[!htbp]
\centering
\small
\setlength{\tabcolsep}{4pt}
\renewcommand{\arraystretch}{1.15}
\caption{ViT-B/16 per-class classification report on the held-out test set.}
\label{tab:2}
\begin{tabular}{>{\raggedright\arraybackslash}p{0.318\linewidth}>{\raggedright\arraybackslash}p{0.146\linewidth}>{\raggedright\arraybackslash}p{0.146\linewidth}>{\raggedright\arraybackslash}p{0.146\linewidth}>{\raggedright\arraybackslash}p{0.146\linewidth}}
\toprule
\nobreak\hspace{0pt}\textbf{Class} & \nobreak\hspace{0pt}\textbf{Precision} & \nobreak\hspace{0pt}\textbf{Recall} & \nobreak\hspace{0pt}\textbf{F1-score} & \nobreak\hspace{0pt}\textbf{Support} \\
\midrule
\nobreak\hspace{0pt}dyed-lifted-polyps & \nobreak\hspace{0pt}0.936 & \nobreak\hspace{0pt}0.880 & \nobreak\hspace{0pt}0.907 & \nobreak\hspace{0pt}50 \\
\nobreak\hspace{0pt}dyed-resection-margins & \nobreak\hspace{0pt}0.889 & \nobreak\hspace{0pt}0.960 & \nobreak\hspace{0pt}0.923 & \nobreak\hspace{0pt}50 \\
\nobreak\hspace{0pt}esophagitis & \nobreak\hspace{0pt}0.857 & \nobreak\hspace{0pt}0.840 & \nobreak\hspace{0pt}0.848 & \nobreak\hspace{0pt}50 \\
\nobreak\hspace{0pt}normal-cecum & \nobreak\hspace{0pt}0.980 & \nobreak\hspace{0pt}1.000 & \nobreak\hspace{0pt}0.990 & \nobreak\hspace{0pt}50 \\
\nobreak\hspace{0pt}normal-pylorus & \nobreak\hspace{0pt}0.962 & \nobreak\hspace{0pt}1.000 & \nobreak\hspace{0pt}0.980 & \nobreak\hspace{0pt}50 \\
\nobreak\hspace{0pt}normal-z-line & \nobreak\hspace{0pt}0.843 & \nobreak\hspace{0pt}0.860 & \nobreak\hspace{0pt}0.851 & \nobreak\hspace{0pt}50 \\
\nobreak\hspace{0pt}polyps & \nobreak\hspace{0pt}0.978 & \nobreak\hspace{0pt}0.880 & \nobreak\hspace{0pt}0.926 & \nobreak\hspace{0pt}50 \\
\nobreak\hspace{0pt}ulcerative-colitis & \nobreak\hspace{0pt}0.941 & \nobreak\hspace{0pt}0.960 & \nobreak\hspace{0pt}0.950 & \nobreak\hspace{0pt}50 \\
\nobreak\hspace{0pt}Macro avg & \nobreak\hspace{0pt}0.923 & \nobreak\hspace{0pt}0.922 & \nobreak\hspace{0pt}0.922 & \nobreak\hspace{0pt}400 \\
\nobreak\hspace{0pt}Accuracy &  &  & \nobreak\hspace{0pt}0.922 & \nobreak\hspace{0pt}400 \\
\bottomrule
\end{tabular}
\end{table}

\begin{table}[!htbp]
\centering
\small
\setlength{\tabcolsep}{4pt}
\renewcommand{\arraystretch}{1.15}
\caption{ResNet50 per-class classification report on the held-out test set.}
\label{tab:3}
\begin{tabular}{>{\raggedright\arraybackslash}p{0.318\linewidth}>{\raggedright\arraybackslash}p{0.146\linewidth}>{\raggedright\arraybackslash}p{0.146\linewidth}>{\raggedright\arraybackslash}p{0.146\linewidth}>{\raggedright\arraybackslash}p{0.146\linewidth}}
\toprule
\nobreak\hspace{0pt}\textbf{Class} & \nobreak\hspace{0pt}\textbf{Precision} & \nobreak\hspace{0pt}\textbf{Recall} & \nobreak\hspace{0pt}\textbf{F1-score} & \nobreak\hspace{0pt}\textbf{Support} \\
\midrule
\nobreak\hspace{0pt}dyed-lifted-polyps & \nobreak\hspace{0pt}1.000 & \nobreak\hspace{0pt}0.860 & \nobreak\hspace{0pt}0.925 & \nobreak\hspace{0pt}50 \\
\nobreak\hspace{0pt}dyed-resection-margins & \nobreak\hspace{0pt}0.877 & \nobreak\hspace{0pt}1.000 & \nobreak\hspace{0pt}0.935 & \nobreak\hspace{0pt}50 \\
\nobreak\hspace{0pt}esophagitis & \nobreak\hspace{0pt}0.875 & \nobreak\hspace{0pt}0.840 & \nobreak\hspace{0pt}0.857 & \nobreak\hspace{0pt}50 \\
\nobreak\hspace{0pt}normal-cecum & \nobreak\hspace{0pt}0.962 & \nobreak\hspace{0pt}1.000 & \nobreak\hspace{0pt}0.980 & \nobreak\hspace{0pt}50 \\
\nobreak\hspace{0pt}normal-pylorus & \nobreak\hspace{0pt}0.980 & \nobreak\hspace{0pt}1.000 & \nobreak\hspace{0pt}0.990 & \nobreak\hspace{0pt}50 \\
\nobreak\hspace{0pt}normal-z-line & \nobreak\hspace{0pt}0.846 & \nobreak\hspace{0pt}0.880 & \nobreak\hspace{0pt}0.863 & \nobreak\hspace{0pt}50 \\
\nobreak\hspace{0pt}polyps & \nobreak\hspace{0pt}0.979 & \nobreak\hspace{0pt}0.940 & \nobreak\hspace{0pt}0.959 & \nobreak\hspace{0pt}50 \\
\nobreak\hspace{0pt}ulcerative-colitis & \nobreak\hspace{0pt}1.000 & \nobreak\hspace{0pt}0.980 & \nobreak\hspace{0pt}0.990 & \nobreak\hspace{0pt}50 \\
\nobreak\hspace{0pt}Macro avg & \nobreak\hspace{0pt}0.940 & \nobreak\hspace{0pt}0.938 & \nobreak\hspace{0pt}0.937 & \nobreak\hspace{0pt}400 \\
\nobreak\hspace{0pt}Accuracy &  &  & \nobreak\hspace{0pt}0.938 & \nobreak\hspace{0pt}400 \\
\bottomrule
\end{tabular}
\end{table}

\subsection{Confusion Matrices and ROC Analysis}

\begin{figure}[!htbp]
\centering
\includegraphics[width=1.0\linewidth,height=0.8\textheight,keepaspectratio]{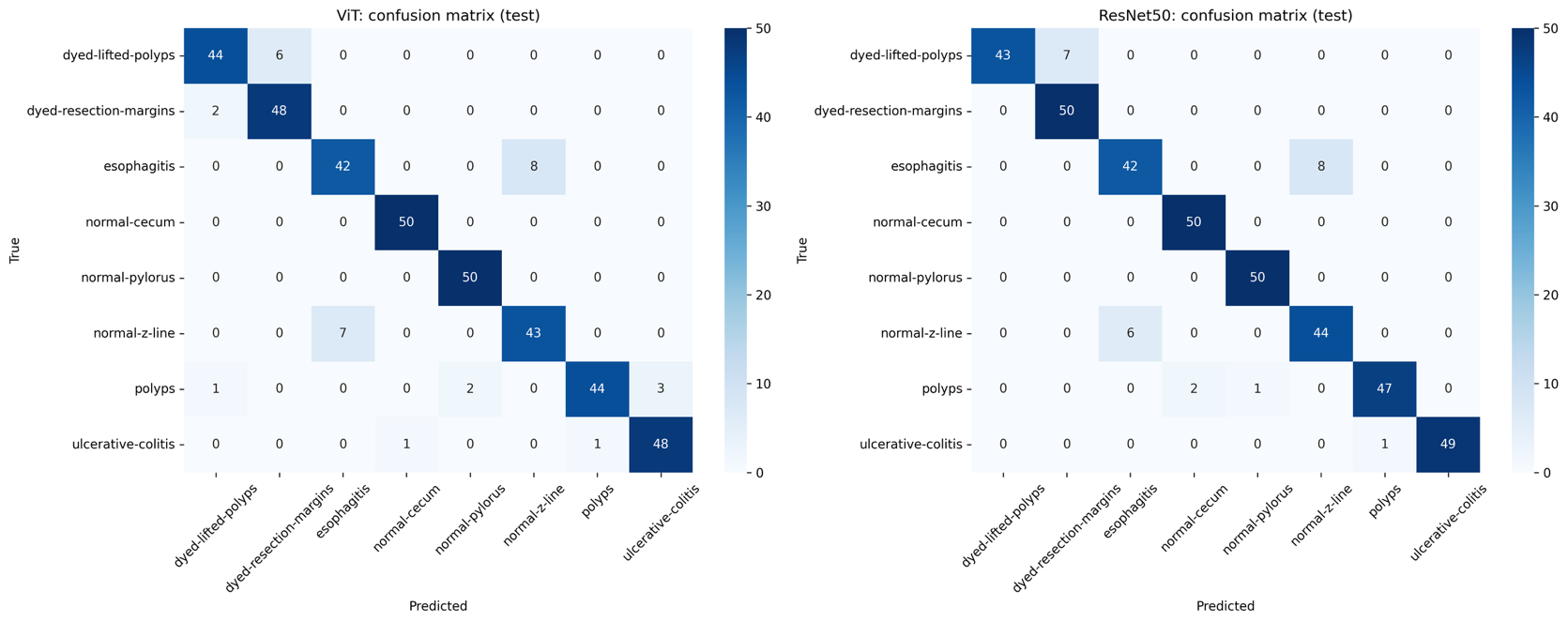}
\caption{Confusion matrices of ViT-B/16 (left) and ResNet50 (right) on the held-out test set (n = 400).}
\label{fig:8}
\end{figure}

\begin{figure}[!htbp]
\centering
\includegraphics[width=0.49\linewidth,height=0.42\textheight,keepaspectratio]{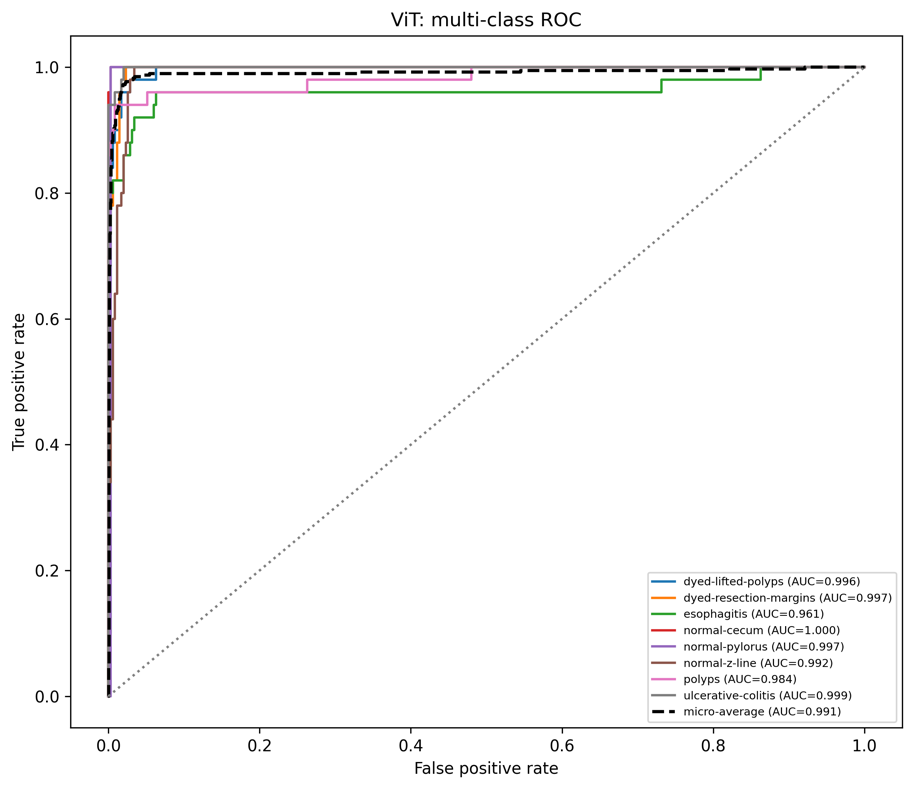}%
\includegraphics[width=0.49\linewidth,height=0.42\textheight,keepaspectratio]{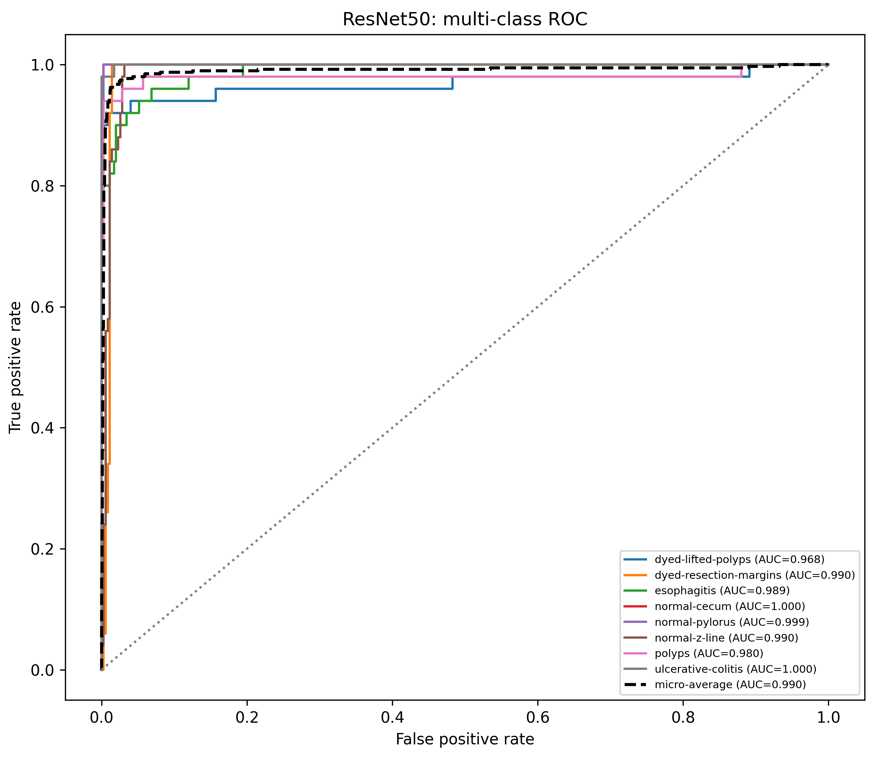}
\caption{Multi-class ROC curves of ViT-B/16 (left) and ResNet50 (right) with per-class and micro-averaged AUC.}
\label{fig:9}
\end{figure}

The confusion matrices are shown in Figure~\ref{fig:8}. Both are strongly diagonal. For both models, most errors fall in two clinically adjacent class pairs: esophagitis versus normal-z-line (14 of 25 ResNet50 errors and 15 of 31 ViT errors) and dyed-lifted-polyps versus dyed-resection-margins (7 and 8 errors). Together these pairs account for 84\% of ResNet50 errors and 74\% of ViT errors on the test split, and for 70\% of the errors of both models in the cross-validation (Section 4.3). Mild esophagitis at the gastro-esophageal junction can resemble a normal Z-line, and endoscopists themselves agree poorly on low-grade esophagitis in this region ($\kappa$ = 0.094 among 91 endoscopists \cite{r49}); the two dyed classes share the same chromoendoscopic appearance and differ mainly in procedural context. The residual errors therefore reflect intrinsically ambiguous distinctions more than architecture-specific failures. The multi-class ROC curves (Figure~\ref{fig:9}) show micro-averaged AUCs of 0.990 (ResNet50) and 0.991 (ViT), with per-class AUCs between 0.96 and 1.00 for both models.

\subsection{Statistical Significance Analysis}

To compare the two models rigorously rather than by point estimates alone, we computed a complete set of statistics on the shared 400-image test set: overall accuracy with 95\% Wilson score confidence intervals; macro-averaged F1 with 95\% non-parametric bootstrap confidence intervals (10,000 resamples); McNemar's paired test (both the continuity-corrected chi-square statistic and the exact binomial p-value) on the discordant predictions; a 95\% paired-bootstrap confidence interval for the difference in accuracy; and per-class F1 with 95\% bootstrap confidence intervals. The overall-performance statistics and the paired comparison are summarized in Table~\ref{tab:4}, the McNemar contingency table is given in Table~\ref{tab:5}, and the per-class confidence intervals in Table~\ref{tab:6}.

\begin{table}[!htbp]
\centering
\small
\setlength{\tabcolsep}{4pt}
\renewcommand{\arraystretch}{1.15}
\caption{Overall test-set performance of ResNet50 and ViT-B/16 with 95\% confidence intervals and the paired statistical comparison (n = 400).}
\label{tab:4}
\begin{tabular}{>{\raggedright\arraybackslash}p{0.383\linewidth}>{\raggedright\arraybackslash}p{0.280\linewidth}>{\raggedright\arraybackslash}p{0.280\linewidth}}
\toprule
\nobreak\hspace{0pt}\textbf{Metric} & \nobreak\hspace{0pt}\textbf{ResNet50 [95\% CI]} & \nobreak\hspace{0pt}\textbf{ViT-B/16 [95\% CI]} \\
\midrule
\nobreak\hspace{0pt}Correct predictions & \nobreak\hspace{0pt}375 / 400 & \nobreak\hspace{0pt}369 / 400 \\
\nobreak\hspace{0pt}Overall accuracy (Wilson CI) & \nobreak\hspace{0pt}93.75\% [90.94, 95.73] & \nobreak\hspace{0pt}92.25\% [89.21, 94.49] \\
\nobreak\hspace{0pt}Macro-averaged F1 (bootstrap CI) & \nobreak\hspace{0pt}0.937 [0.913, 0.960] & \nobreak\hspace{0pt}0.922 [0.895, 0.947] \\
\nobreak\hspace{0pt}Accuracy difference (R50 $-$ ViT), 95\% paired bootstrap CI & \nobreak\hspace{0pt}+1.50 pp [$-$0.50, +3.50] & \nobreak\hspace{0pt}(paired) \\
\nobreak\hspace{0pt}90\% CI for the difference / TOST p ($\pm$3 pp) & \nobreak\hspace{0pt}[0.00, +3.25] / 0.066 & \nobreak\hspace{0pt}(paired) \\
\nobreak\hspace{0pt}McNemar exact p / $\chi$$^{2}$ (continuity-corrected) & \nobreak\hspace{0pt}p = 0.21 & \nobreak\hspace{0pt}$\chi$$^{2}$ = 1.56 \\
\bottomrule
\end{tabular}
\end{table}

Because the accuracy difference between the two models is small (1.50 percentage points), we tested both whether it is significant and whether the models are equivalent. Of the 400 test images, both models were correct on 364, both were wrong on 20, ResNet50 alone was correct on 11 and ViT alone on 5 (Table~\ref{tab:5}). McNemar's test on the 16 discordant pairs gave a continuity-corrected chi-square of 1.56 and an exact p-value of 0.21, and the 95\% paired-bootstrap interval for the difference (ResNet50 minus ViT) was [$-$0.50\%, +3.50\%].

\begin{table}[!htbp]
\centering
\small
\setlength{\tabcolsep}{4pt}
\renewcommand{\arraystretch}{1.15}
\caption{McNemar's paired contingency table comparing ResNet50 and ViT on the shared test set.}
\label{tab:5}
\begin{tabular}{>{\raggedright\arraybackslash}p{0.340\linewidth}>{\raggedright\arraybackslash}p{0.301\linewidth}>{\raggedright\arraybackslash}p{0.301\linewidth}}
\toprule
\nobreak\hspace{0pt}\textbf{Models} & \nobreak\hspace{0pt}\textbf{ViT correct} & \nobreak\hspace{0pt}\textbf{ViT incorrect} \\
\midrule
\nobreak\hspace{0pt}ResNet50 correct & \nobreak\hspace{0pt}364 & \nobreak\hspace{0pt}11 \\
\nobreak\hspace{0pt}ResNet50 incorrect & \nobreak\hspace{0pt}5 & \nobreak\hspace{0pt}20 \\
\nobreak\hspace{0pt}McNemar's test & \nobreak\hspace{0pt}$\chi$$^{2}$ (cc) = 1.56 & \nobreak\hspace{0pt}exact p = 0.21 \\
\bottomrule
\end{tabular}
\end{table}

The difference is therefore not statistically significant, but a non-significant result does not establish equivalence. With 400 test images the 90\% confidence interval for the difference was [0.00, +3.25] percentage points, so equivalence within $\pm$3 points could not be shown on this split (TOST p = 0.066; the smallest margin at which equivalence would be declared is 3.25 points). At the class level (Table~\ref{tab:6}) the 95\% bootstrap intervals for per-class F1 overlap for every class; overlapping intervals are not a formal test, and with 50 test images per class the split is not powered to detect or exclude class-level differences. Removing the 37 test images that had a candidate near-duplicate in the training or validation data left the result unchanged (ResNet50 94.21\% vs. ViT 92.56\%, n = 363, McNemar p = 0.18).

\begin{table}[!htbp]
\centering
\small
\setlength{\tabcolsep}{4pt}
\renewcommand{\arraystretch}{1.15}
\caption{Per-class F1-score with 95\% bootstrap confidence intervals for both models on the test set.}
\label{tab:6}
\begin{tabular}{>{\raggedright\arraybackslash}p{0.340\linewidth}>{\raggedright\arraybackslash}p{0.301\linewidth}>{\raggedright\arraybackslash}p{0.301\linewidth}}
\toprule
\nobreak\hspace{0pt}\textbf{Class} & \nobreak\hspace{0pt}\textbf{ResNet50 F1 [95\% CI]} & \nobreak\hspace{0pt}\textbf{ViT-B/16 F1 [95\% CI]} \\
\midrule
\nobreak\hspace{0pt}dyed-lifted-polyps & \nobreak\hspace{0pt}0.92 [0.86, 0.98] & \nobreak\hspace{0pt}0.91 [0.84, 0.96] \\
\nobreak\hspace{0pt}dyed-resection-margins & \nobreak\hspace{0pt}0.93 [0.88, 0.98] & \nobreak\hspace{0pt}0.92 [0.86, 0.97] \\
\nobreak\hspace{0pt}esophagitis & \nobreak\hspace{0pt}0.86 [0.78, 0.93] & \nobreak\hspace{0pt}0.85 [0.76, 0.92] \\
\nobreak\hspace{0pt}normal-cecum & \nobreak\hspace{0pt}0.98 [0.95, 1.00] & \nobreak\hspace{0pt}0.99 [0.97, 1.00] \\
\nobreak\hspace{0pt}normal-pylorus & \nobreak\hspace{0pt}0.99 [0.97, 1.00] & \nobreak\hspace{0pt}0.98 [0.95, 1.00] \\
\nobreak\hspace{0pt}normal-z-line & \nobreak\hspace{0pt}0.86 [0.78, 0.93] & \nobreak\hspace{0pt}0.85 [0.77, 0.92] \\
\nobreak\hspace{0pt}polyps & \nobreak\hspace{0pt}0.96 [0.91, 0.99] & \nobreak\hspace{0pt}0.93 [0.86, 0.97] \\
\nobreak\hspace{0pt}ulcerative-colitis & \nobreak\hspace{0pt}0.99 [0.97, 1.00] & \nobreak\hspace{0pt}0.95 [0.90, 0.99] \\
\bottomrule
\end{tabular}
\end{table}

\begin{figure}[!htbp]
\centering
\includegraphics[width=1.0\linewidth,height=0.8\textheight,keepaspectratio]{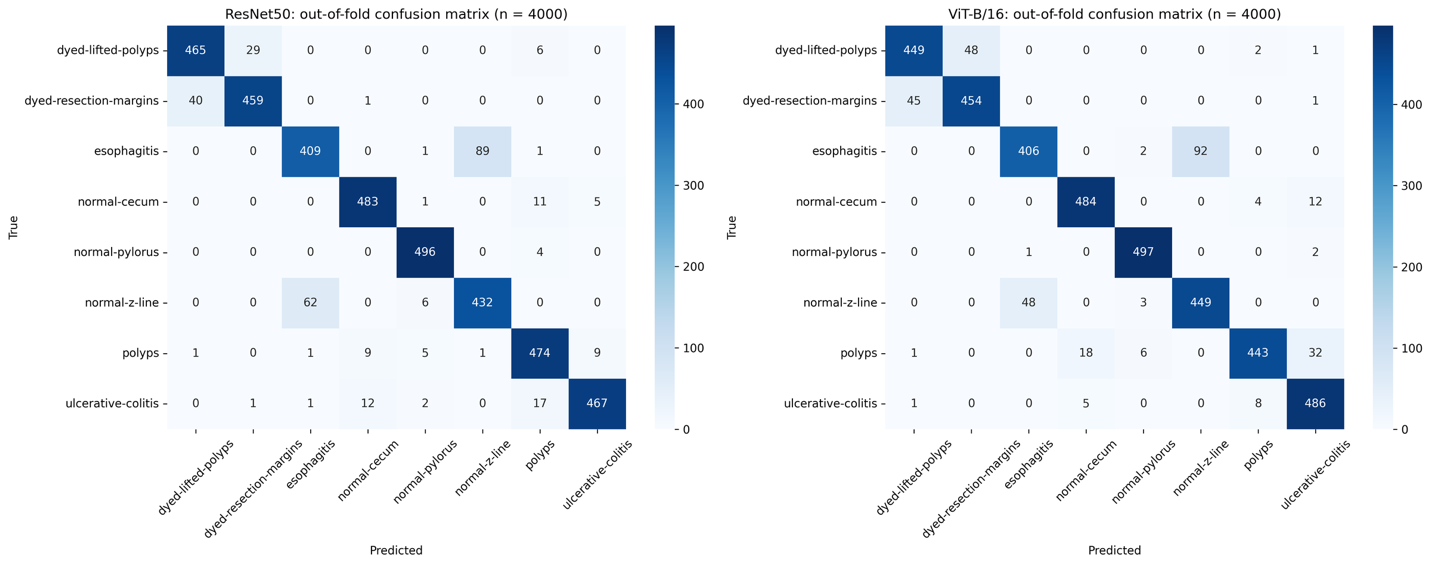}
\caption{Out-of-fold confusion matrices of ResNet50 (left) and ViT-B/16 (right) from the 5-fold cross-validation (n = 4,000).}
\label{fig:10}
\end{figure}

\textbf{Cross-validation (confirmatory analysis).} Stratified group 5-fold cross-validation (Table~\ref{tab:7}) gave fold-wise accuracies of 92.13 $\pm$ 0.76\% for ResNet50 and 91.70 $\pm$ 0.72\% for ViT, and pooled out-of-fold accuracies of 92.12\% [91.25, 92.92] and 91.70\% [90.80, 92.52] over all 4,000 images. The pooled difference was +0.43 percentage points (90\% CI $-$0.33 to +1.18; McNemar p = 0.39 on 178 vs. 161 discordant images), well inside the $\pm$3-point margin (TOST p \textless{} 0.001); equivalence would hold for any margin of 1.2 points or more. The mean fold-wise difference was also +0.43 points (90\% t-interval $-$0.01 to +0.88). Cross-validated accuracy was about 1.5 points lower than on the primary split for both models, consistent with the smaller training sets and with the group-aware folds, which keep candidate near-duplicates out of the test folds. The out-of-fold confusion matrices (Figure~\ref{fig:10}) show the same error structure as the primary split.

\begin{table}[!htbp]
\centering
\small
\setlength{\tabcolsep}{4pt}
\renewcommand{\arraystretch}{1.15}
\caption{Stratified group 5-fold cross-validation of ResNet50 and ViT-B/16 (n = 4,000 pooled out-of-fold predictions per model).}
\label{tab:7}
\begin{tabular}{>{\raggedright\arraybackslash}p{0.447\linewidth}>{\raggedright\arraybackslash}p{0.248\linewidth}>{\raggedright\arraybackslash}p{0.248\linewidth}}
\toprule
\nobreak\hspace{0pt}\textbf{Metric} & \nobreak\hspace{0pt}\textbf{ResNet50} & \nobreak\hspace{0pt}\textbf{ViT-B/16} \\
\midrule
\nobreak\hspace{0pt}Fold-wise accuracy, mean $\pm$ SD (\%) & \nobreak\hspace{0pt}92.13 $\pm$ 0.76 & \nobreak\hspace{0pt}91.70 $\pm$ 0.72 \\
\nobreak\hspace{0pt}Fold-wise macro-F1, mean $\pm$ SD ($\times$100) & \nobreak\hspace{0pt}92.10 $\pm$ 0.78 & \nobreak\hspace{0pt}91.67 $\pm$ 0.77 \\
\nobreak\hspace{0pt}Pooled out-of-fold accuracy [95\% Wilson CI] (\%) & \nobreak\hspace{0pt}92.12 [91.25, 92.92] & \nobreak\hspace{0pt}91.70 [90.80, 92.52] \\
\nobreak\hspace{0pt}Pooled $\Delta$ accuracy (R50 $-$ ViT) [90\% bootstrap CI] (pp) & \nobreak\hspace{0pt}+0.43 [$-$0.33, +1.18] & \nobreak\hspace{0pt}(paired) \\
\nobreak\hspace{0pt}McNemar exact p (pooled; discordant 178 / 161) & \nobreak\hspace{0pt}0.385 & \nobreak\hspace{0pt}(paired) \\
\nobreak\hspace{0pt}TOST p ($\pm$3 pp) / smallest equivalence margin (pp) & \nobreak\hspace{0pt}\textless{} 0.001 / 1.18 & \nobreak\hspace{0pt}(paired) \\
\nobreak\hspace{0pt}Fold-wise $\Delta$, mean [90\% t-CI] (pp) & \nobreak\hspace{0pt}+0.43 [$-$0.01, +0.88] & \nobreak\hspace{0pt}(paired) \\
\nobreak\hspace{0pt}Best epoch, mean $\pm$ SD & \nobreak\hspace{0pt}6.2 $\pm$ 1.5 & \nobreak\hspace{0pt}8.6 $\pm$ 7.5 \\
\nobreak\hspace{0pt}Epoch time, mean (s) & \nobreak\hspace{0pt}35 & \nobreak\hspace{0pt}106 \\
\nobreak\hspace{0pt}Time to best checkpoint, mean (min) & \nobreak\hspace{0pt}3.8 & \nobreak\hspace{0pt}15.8 \\
\bottomrule
\end{tabular}
\end{table}

\subsection{Computational Efficiency}

The models differ markedly in computational cost (Table~\ref{tab:8} summarizes all metrics; Table~\ref{tab:9} gives the standardized benchmark). ResNet50 has 23.52 million parameters and requires 4.13 GMACs per image, whereas ViT-B/16 has 85.80 million parameters and requires 17.61 GMACs, roughly 3.6 times more parameters and 4.3 times more computation. On the T4 GPU, ResNet50's single-image latency was 5.4 ms versus 13.1 ms for ViT, and its throughput at batch size 32 was 337 versus 88 images per second in FP32 (763 vs. 289 in FP16). On the CPU, single-image latency was 69 ms versus 177 ms and throughput 10.8 versus 5.2 images per second. Peak GPU memory at batch size 32 was 436 MB versus 595 MB. The efficiency gap extends to training: in cross-validation ResNet50 needed 35 s per epoch and reached its best checkpoint after 3.8 min on average, against 106 s and 15.8 min for ViT.

\begin{table}[!htbp]
\centering
\small
\setlength{\tabcolsep}{4pt}
\renewcommand{\arraystretch}{1.15}
\caption{Summary comparison of ResNet50, ViT-B/16 and MobileViT-S across accuracy, efficiency, calibration and robustness.}
\label{tab:8}
\begin{tabular}{>{\raggedright\arraybackslash}p{0.383\linewidth}>{\raggedright\arraybackslash}p{0.180\linewidth}>{\raggedright\arraybackslash}p{0.180\linewidth}>{\raggedright\arraybackslash}p{0.180\linewidth}}
\toprule
\nobreak\hspace{0pt}\textbf{Metric} & \nobreak\hspace{0pt}\textbf{ResNet50} & \nobreak\hspace{0pt}\textbf{ViT-B/16} & \nobreak\hspace{0pt}\textbf{Mobile\-ViT-S} \\
\midrule
\nobreak\hspace{0pt}Test accuracy, primary split (\%) & \nobreak\hspace{0pt}93.75 & \nobreak\hspace{0pt}92.25 & \nobreak\hspace{0pt}95.50 \\
\nobreak\hspace{0pt}Macro-F1, primary split & \nobreak\hspace{0pt}0.937 & \nobreak\hspace{0pt}0.922 & \nobreak\hspace{0pt}0.955 \\
\nobreak\hspace{0pt}Cross-validated accuracy (\%) & \nobreak\hspace{0pt}92.12 & \nobreak\hspace{0pt}91.70 & \nobreak\hspace{0pt}--- \\
\nobreak\hspace{0pt}External accuracy, GastroVision (\%) & \nobreak\hspace{0pt}78.08 & \nobreak\hspace{0pt}79.63 & \nobreak\hspace{0pt}78.33 \\
\nobreak\hspace{0pt}Parameters (M) / MACs (G) & \nobreak\hspace{0pt}23.52 / 4.13 & \nobreak\hspace{0pt}85.80 / 17.61 & \nobreak\hspace{0pt}4.94 / 1.24 \\
\nobreak\hspace{0pt}GPU throughput, batch 32, FP32 (img/s) & \nobreak\hspace{0pt}336.5 & \nobreak\hspace{0pt}87.9 & \nobreak\hspace{0pt}444.4 \\
\nobreak\hspace{0pt}CPU latency, batch 1 (ms) & \nobreak\hspace{0pt}69.4 & \nobreak\hspace{0pt}176.6 & \nobreak\hspace{0pt}53.5 \\
\nobreak\hspace{0pt}ECE, 15 bins: primary split / cross-validation & \nobreak\hspace{0pt}0.025 / 0.021 & \nobreak\hspace{0pt}0.039 / 0.023 & \nobreak\hspace{0pt}0.032 / --- \\
\nobreak\hspace{0pt}Accuracy change when the ScopeGuide inset is occluded (pp) & \nobreak\hspace{0pt}$-$1.3 (p = 0.69) & \nobreak\hspace{0pt}$-$4.0 (p = 0.15) & \nobreak\hspace{0pt}$-$9.4 (p \textless{} 0.001) \\
\bottomrule
\end{tabular}
\end{table}

\begin{figure}[!htbp]
\centering
\includegraphics[width=0.97\linewidth,height=0.8\textheight,keepaspectratio]{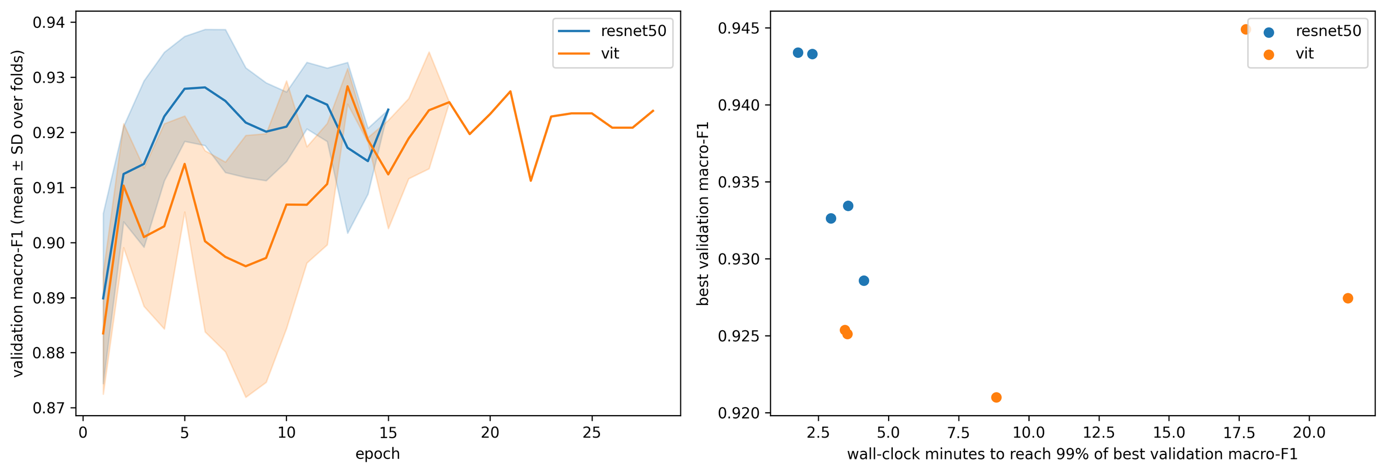}
\caption{Convergence across the five cross-validation folds. Left: validation macro-F1 per epoch (mean $\pm$ SD); right: wall-clock minutes needed to reach 99\% of each fold's best validation macro-F1.}
\label{fig:11}
\end{figure}

\textbf{Convergence:} Across the five folds, ResNet50 reached its best validation macro-F1 at epoch 6.2 $\pm$ 1.5 and 99\% of that value after 5.0 $\pm$ 1.6 epochs (2.9 $\pm$ 0.9 min), whereas ViT's best epoch varied widely (8.6 $\pm$ 7.5; range 2--21) and it needed 6.2 $\pm$ 4.6 epochs (11.0 $\pm$ 8.2 min) to reach 99\% of its best value (Figure~\ref{fig:11}). The later best epoch of ViT on the primary split (20 vs. 8) is therefore not a stable property of the architecture: ViT's validation curve is noisier, so its best checkpoint occurs at an unpredictable epoch, and each epoch costs three times as much. The best validation macro-F1 values were similar (0.94 $\pm$ 0.01 vs. 0.93 $\pm$ 0.01).

\textbf{MobileViT-S:} MobileViT-S has 4.94 million parameters and 1.24 GMACs (a ptflops estimate that may undercount its attention operations). It had the lowest CPU latency (53.5 ms), the highest CPU throughput (13.1 images per second) and the highest FP32 GPU throughput (444 images per second), although its single-image GPU latency (7.6 ms) was higher than ResNet50's. At batch size 1, GPU latency is dominated by kernel-launch overhead, so FP16 did not reduce latency for ResNet50 or MobileViT-S. MobileViT-S's large early feature maps made it the most memory-hungry model on the CPU at batch size 32 (711 MB, against 445 MB for ResNet50 and 343 MB for ViT).

\begin{table}[!htbp]
\centering
\small
\setlength{\tabcolsep}{4pt}
\renewcommand{\arraystretch}{1.15}
\caption{Standardized inference benchmark and training cost (Tesla T4 GPU; Intel Xeon 2.0 GHz CPU, 4 threads; PyTorch 2.10). Latencies are medians, with the interquartile range in brackets, over 1,000 timed GPU or 150 timed CPU iterations. Training cost is the cross-validation mean for ResNet50 and ViT-B/16 and the primary run for MobileViT-S.}
\label{tab:9}
\begin{tabular}{>{\raggedright\arraybackslash}p{0.383\linewidth}>{\raggedright\arraybackslash}p{0.180\linewidth}>{\raggedright\arraybackslash}p{0.180\linewidth}>{\raggedright\arraybackslash}p{0.180\linewidth}}
\toprule
\nobreak\hspace{0pt}\textbf{Metric} & \nobreak\hspace{0pt}\textbf{ResNet50} & \nobreak\hspace{0pt}\textbf{ViT-B/16} & \nobreak\hspace{0pt}\textbf{Mobile\-ViT-S} \\
\midrule
\nobreak\hspace{0pt}Parameters (M) & \nobreak\hspace{0pt}23.52 & \nobreak\hspace{0pt}85.80 & \nobreak\hspace{0pt}4.94 \\
\nobreak\hspace{0pt}MACs (G) & \nobreak\hspace{0pt}4.13 & \nobreak\hspace{0pt}17.61 & \nobreak\hspace{0pt}1.24 \\
\nobreak\hspace{0pt}GPU latency, batch 1, FP32 (ms) & \nobreak\hspace{0pt}5.4 [5.3--5.5] & \nobreak\hspace{0pt}13.1 [13.1--13.4] & \nobreak\hspace{0pt}7.6 [7.5--7.8] \\
\nobreak\hspace{0pt}GPU latency, batch 1, FP16 (ms) & \nobreak\hspace{0pt}5.8 [5.7--5.9] & \nobreak\hspace{0pt}5.9 [5.8--6.1] & \nobreak\hspace{0pt}10.0 [9.9--10.2] \\
\nobreak\hspace{0pt}GPU throughput, batch 32, FP32 (img/s) & \nobreak\hspace{0pt}336.5 & \nobreak\hspace{0pt}87.9 & \nobreak\hspace{0pt}444.4 \\
\nobreak\hspace{0pt}GPU throughput, batch 32, FP16 (img/s) & \nobreak\hspace{0pt}762.8 & \nobreak\hspace{0pt}288.6 & \nobreak\hspace{0pt}662.5 \\
\nobreak\hspace{0pt}Peak GPU memory, batch 32, FP32 (MB) & \nobreak\hspace{0pt}436 & \nobreak\hspace{0pt}595 & \nobreak\hspace{0pt}512 \\
\nobreak\hspace{0pt}CPU latency, batch 1 (ms) & \nobreak\hspace{0pt}69.4 [67.7--71.7] & \nobreak\hspace{0pt}176.6 [173.3--182.5] & \nobreak\hspace{0pt}53.5 [52.8--55.0] \\
\nobreak\hspace{0pt}CPU throughput, batch 32 (img/s) & \nobreak\hspace{0pt}10.77 & \nobreak\hspace{0pt}5.22 & \nobreak\hspace{0pt}13.10 \\
\nobreak\hspace{0pt}CPU inference memory, batch 1 / 32 (MB) & \nobreak\hspace{0pt}43 / 445 & \nobreak\hspace{0pt}37 / 343 & \nobreak\hspace{0pt}44 / 711 \\
\nobreak\hspace{0pt}Training time per epoch (s) & \nobreak\hspace{0pt}35 & \nobreak\hspace{0pt}106 & \nobreak\hspace{0pt}30 \\
\nobreak\hspace{0pt}Time to best checkpoint (min) & \nobreak\hspace{0pt}3.8 & \nobreak\hspace{0pt}15.8 & \nobreak\hspace{0pt}6.2 \\
\bottomrule
\end{tabular}
\end{table}

\subsection{Calibration}

\begin{figure}[!htbp]
\centering
\includegraphics[width=0.97\linewidth,height=0.8\textheight,keepaspectratio]{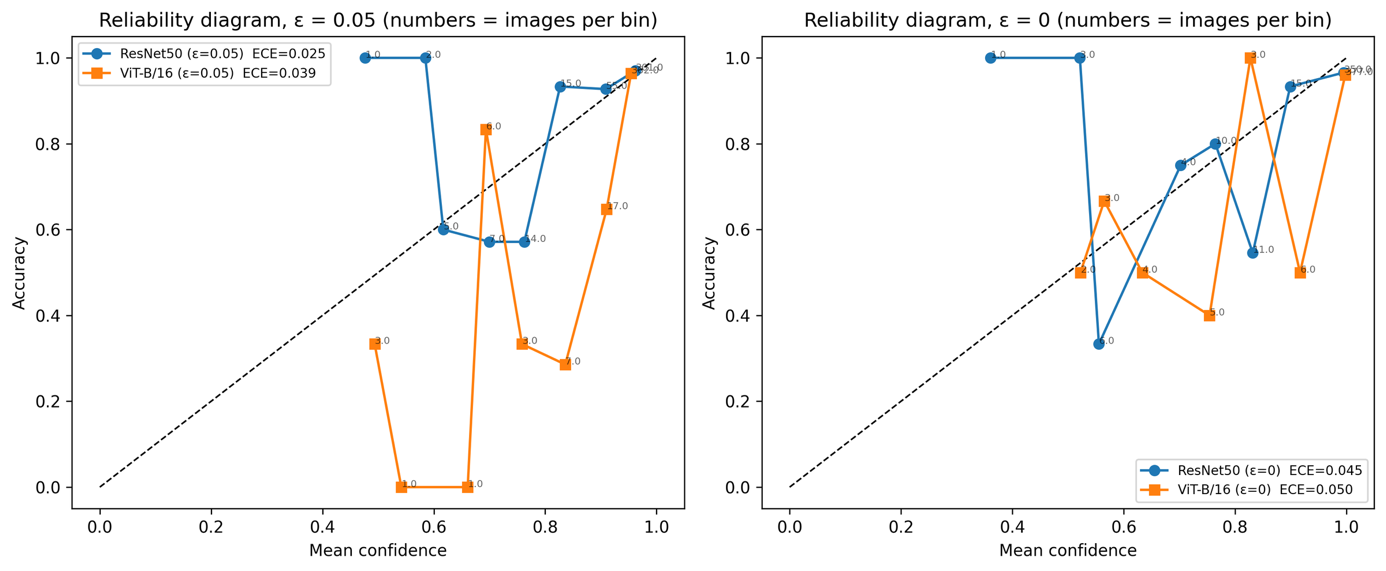}
\caption{Reliability diagrams (15 equal-width bins; numbers are images per bin) for ResNet50 and ViT-B/16 trained with (left) and without (right) label smoothing.}
\label{fig:12}
\end{figure}

\begin{figure}[!htbp]
\centering
\includegraphics[width=1.0\linewidth,height=0.8\textheight,keepaspectratio]{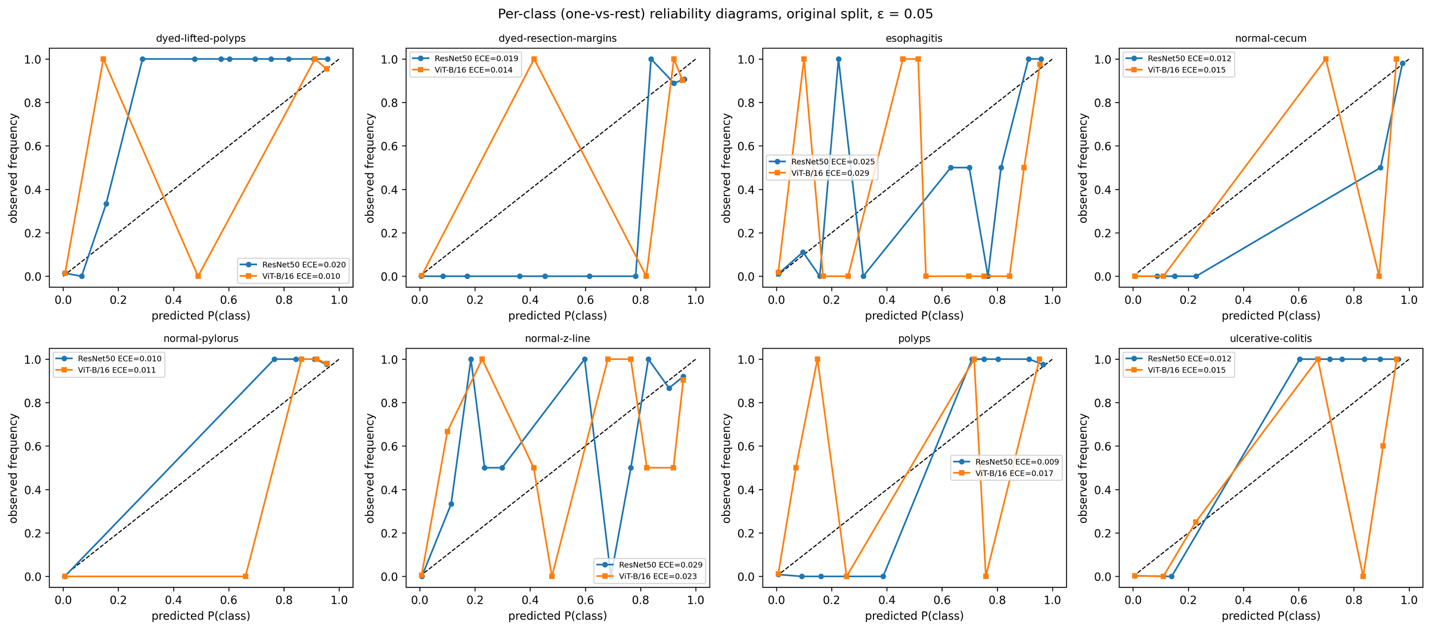}
\caption{Per-class (one-vs-rest) reliability diagrams of ResNet50 and ViT-B/16 on the primary split ($\varepsilon$ = 0.05), with per-class ECE (15 equal-width bins) in the legends.}
\label{fig:13}
\end{figure}

Calibration results are reported in Table~\ref{tab:10} and Figure~\ref{fig:12}. On the primary split, ResNet50 and ViT-B/16 had ECEs of 0.025 [0.017, 0.050] and 0.039 [0.023, 0.065] with 15 equal-width bins. The ranking, however, depended on the binning: with 10 equal-width bins ViT appeared better calibrated (0.020 vs. 0.025), whereas with 15 equal-mass bins ResNet50 was better (0.032 vs. 0.060; paired difference $-$0.027 [$-$0.047, $-$0.001]). Class-wise ECE (0.017 for both), the Brier score (0.107 vs. 0.127) and NLL (0.248 vs. 0.291) did not differ significantly. In cross-validation, where 4,000 predictions give much tighter intervals, the models were indistinguishable (ECE 0.021 vs. 0.023, paired difference $-$0.001 [$-$0.011, +0.007]; Brier 0.122 vs. 0.128; NLL 0.265 vs. 0.272). Per-class ECE was highest for esophagitis and normal-z-line (0.023--0.029 on the primary split) and at most 0.017 for every class in cross-validation (Table~\ref{tab:11} and Figure~\ref{fig:13}). The primary-split ECE ranking is therefore fragile; the reliability diagrams show why, as almost all predictions fall in the top confidence bin and the remaining bins hold only a handful of images each (Figure~\ref{fig:12}).

\textbf{Effect of label smoothing:} Without label smoothing ($\varepsilon$ = 0), both models reached 93.75\% on the primary split (8 vs. 8 discordant images; equivalent within $\pm$3 points, TOST p = 0.001). ECE increased for both (ResNet50 0.025 $\rightarrow$ 0.045, ViT 0.039 $\rightarrow$ 0.050), consistent with the regularizing effect of smoothing on over-confidence \cite{r41}. The ResNet50 $-$ ViT ECE difference was $-$0.014 [$-$0.036, +0.015] with smoothing and $-$0.005 [$-$0.024, +0.016] without, so label smoothing did not artificially narrow the calibration gap between the architectures; in both settings the gap was small and not significant.

\begin{table}[!htbp]
\centering
\scriptsize
\setlength{\tabcolsep}{4pt}
\renewcommand{\arraystretch}{1.15}
\caption{Calibration metrics with 95\% bootstrap confidence intervals: primary split with ($\varepsilon$ = 0.05) and without ($\varepsilon$ = 0) label smoothing, MobileViT-S, and cross-validation (CV) out-of-fold predictions.}
\label{tab:10}
\begin{tabular}{>{\raggedright\arraybackslash}p{0.149\linewidth}>{\raggedright\arraybackslash}p{0.100\linewidth}>{\raggedright\arraybackslash}p{0.100\linewidth}>{\raggedright\arraybackslash}p{0.100\linewidth}>{\raggedright\arraybackslash}p{0.100\linewidth}>{\raggedright\arraybackslash}p{0.100\linewidth}>{\raggedright\arraybackslash}p{0.100\linewidth}>{\raggedright\arraybackslash}p{0.100\linewidth}}
\toprule
\nobreak\hspace{0pt}\textbf{Metric} & \nobreak\hspace{0pt}\textbf{ResNet50 $\varepsilon$ = 0.05} & \nobreak\hspace{0pt}\textbf{ViT-B/16 $\varepsilon$ = 0.05} & \nobreak\hspace{0pt}\textbf{ResNet50 $\varepsilon$ = 0} & \nobreak\hspace{0pt}\textbf{ViT-B/16 $\varepsilon$ = 0} & \nobreak\hspace{0pt}\textbf{Mobile\-ViT-S} & \nobreak\hspace{0pt}\textbf{ResNet50 CV} & \nobreak\hspace{0pt}\textbf{ViT-B/16 CV} \\
\midrule
\nobreak\hspace{0pt}Accuracy & \nobreak\hspace{0pt}0.938 [0.912, 0.960] & \nobreak\hspace{0pt}0.922 [0.895, 0.948] & \nobreak\hspace{0pt}0.938 [0.912, 0.960] & \nobreak\hspace{0pt}0.938 [0.912, 0.960] & \nobreak\hspace{0pt}0.955 [0.935, 0.975] & \nobreak\hspace{0pt}0.921 [0.913, 0.929] & \nobreak\hspace{0pt}0.917 [0.908, 0.926] \\
\nobreak\hspace{0pt}ECE, 15 equal-width bins & \nobreak\hspace{0pt}0.025 [0.017, 0.050] & \nobreak\hspace{0pt}0.039 [0.023, 0.065] & \nobreak\hspace{0pt}0.045 [0.030, 0.071] & \nobreak\hspace{0pt}0.050 [0.033, 0.075] & \nobreak\hspace{0pt}0.032 [0.023, 0.054] & \nobreak\hspace{0pt}0.021 [0.017, 0.030] & \nobreak\hspace{0pt}0.023 [0.019, 0.032] \\
\nobreak\hspace{0pt}ECE, 10 equal-width bins & \nobreak\hspace{0pt}0.025 [0.014, 0.048] & \nobreak\hspace{0pt}0.020 [0.015, 0.049] & \nobreak\hspace{0pt}0.039 [0.023, 0.064] & \nobreak\hspace{0pt}0.050 [0.032, 0.074] & \nobreak\hspace{0pt}0.031 [0.019, 0.052] & \nobreak\hspace{0pt}0.017 [0.013, 0.026] & \nobreak\hspace{0pt}0.018 [0.013, 0.026] \\
\nobreak\hspace{0pt}ECE, 15 equal-mass bins & \nobreak\hspace{0pt}0.032 [0.030, 0.061] & \nobreak\hspace{0pt}0.060 [0.047, 0.089] & \nobreak\hspace{0pt}0.030 [0.018, 0.057] & \nobreak\hspace{0pt}0.046 [0.025, 0.069] & \nobreak\hspace{0pt}0.042 [0.035, 0.066] & \nobreak\hspace{0pt}0.016 [0.014, 0.026] & \nobreak\hspace{0pt}0.019 [0.017, 0.029] \\
\nobreak\hspace{0pt}Class-wise ECE & \nobreak\hspace{0pt}0.017 [0.015, 0.024] & \nobreak\hspace{0pt}0.017 [0.014, 0.023] & \nobreak\hspace{0pt}0.015 [0.011, 0.021] & \nobreak\hspace{0pt}0.015 [0.010, 0.021] & \nobreak\hspace{0pt}0.013 [0.013, 0.019] & \nobreak\hspace{0pt}0.013 [0.013, 0.015] & \nobreak\hspace{0pt}0.011 [0.011, 0.014] \\
\nobreak\hspace{0pt}Brier score & \nobreak\hspace{0pt}0.107 [0.071, 0.145] & \nobreak\hspace{0pt}0.127 [0.086, 0.170] & \nobreak\hspace{0pt}0.105 [0.069, 0.144] & \nobreak\hspace{0pt}0.107 [0.068, 0.149] & \nobreak\hspace{0pt}0.080 [0.050, 0.113] & \nobreak\hspace{0pt}0.122 [0.110, 0.134] & \nobreak\hspace{0pt}0.128 [0.116, 0.140] \\
\nobreak\hspace{0pt}NLL & \nobreak\hspace{0pt}0.248 [0.174, 0.329] & \nobreak\hspace{0pt}0.291 [0.207, 0.380] & \nobreak\hspace{0pt}0.226 [0.139, 0.322] & \nobreak\hspace{0pt}0.268 [0.161, 0.387] & \nobreak\hspace{0pt}0.194 [0.136, 0.260] & \nobreak\hspace{0pt}0.265 [0.243, 0.288] & \nobreak\hspace{0pt}0.272 [0.250, 0.294] \\
\bottomrule
\end{tabular}
\end{table}

\begin{table}[!htbp]
\centering
\scriptsize
\setlength{\tabcolsep}{4pt}
\renewcommand{\arraystretch}{1.15}
\caption{Per-class (one-vs-rest) expected calibration error (15 equal-width bins) with 95\% bootstrap confidence intervals. $\varepsilon$ = label smoothing; CV = cross-validation out-of-fold predictions (n = 4,000).}
\label{tab:11}
\begin{tabular}{>{\raggedright\arraybackslash}p{0.149\linewidth}>{\raggedright\arraybackslash}p{0.100\linewidth}>{\raggedright\arraybackslash}p{0.100\linewidth}>{\raggedright\arraybackslash}p{0.100\linewidth}>{\raggedright\arraybackslash}p{0.100\linewidth}>{\raggedright\arraybackslash}p{0.100\linewidth}>{\raggedright\arraybackslash}p{0.100\linewidth}>{\raggedright\arraybackslash}p{0.100\linewidth}}
\toprule
\nobreak\hspace{0pt}\textbf{Class} & \nobreak\hspace{0pt}\textbf{ResNet50 $\varepsilon$ = 0.05} & \nobreak\hspace{0pt}\textbf{ViT-B/16 $\varepsilon$ = 0.05} & \nobreak\hspace{0pt}\textbf{ResNet50 $\varepsilon$ = 0} & \nobreak\hspace{0pt}\textbf{ViT-B/16 $\varepsilon$ = 0} & \nobreak\hspace{0pt}\textbf{Mobile\-ViT-S} & \nobreak\hspace{0pt}\textbf{ResNet50 CV} & \nobreak\hspace{0pt}\textbf{ViT-B/16 CV} \\
\midrule
\nobreak\hspace{0pt}dyed-lifted-polyps & \nobreak\hspace{0pt}0.020 [0.011, 0.035] & \nobreak\hspace{0pt}0.010 [0.004, 0.025] & \nobreak\hspace{0pt}0.017 [0.008, 0.029] & \nobreak\hspace{0pt}0.015 [0.005, 0.028] & \nobreak\hspace{0pt}0.012 [0.009, 0.021] & \nobreak\hspace{0pt}0.010 [0.009, 0.015] & \nobreak\hspace{0pt}0.008 [0.006, 0.013] \\
\nobreak\hspace{0pt}dyed-resection-margins & \nobreak\hspace{0pt}0.019 [0.012, 0.033] & \nobreak\hspace{0pt}0.014 [0.005, 0.026] & \nobreak\hspace{0pt}0.016 [0.007, 0.029] & \nobreak\hspace{0pt}0.013 [0.005, 0.024] & \nobreak\hspace{0pt}0.015 [0.011, 0.024] & \nobreak\hspace{0pt}0.010 [0.008, 0.014] & \nobreak\hspace{0pt}0.010 [0.008, 0.015] \\
\nobreak\hspace{0pt}esophagitis & \nobreak\hspace{0pt}0.025 [0.018, 0.044] & \nobreak\hspace{0pt}0.029 [0.015, 0.047] & \nobreak\hspace{0pt}0.028 [0.020, 0.048] & \nobreak\hspace{0pt}0.031 [0.016, 0.048] & \nobreak\hspace{0pt}0.019 [0.013, 0.036] & \nobreak\hspace{0pt}0.014 [0.011, 0.020] & \nobreak\hspace{0pt}0.009 [0.007, 0.016] \\
\nobreak\hspace{0pt}normal-cecum & \nobreak\hspace{0pt}0.012 [0.009, 0.019] & \nobreak\hspace{0pt}0.015 [0.011, 0.021] & \nobreak\hspace{0pt}0.006 [0.002, 0.014] & \nobreak\hspace{0pt}0.003 [0.000, 0.008] & \nobreak\hspace{0pt}0.016 [0.011, 0.024] & \nobreak\hspace{0pt}0.011 [0.010, 0.015] & \nobreak\hspace{0pt}0.012 [0.011, 0.015] \\
\nobreak\hspace{0pt}normal-pylorus & \nobreak\hspace{0pt}0.010 [0.007, 0.014] & \nobreak\hspace{0pt}0.011 [0.007, 0.017] & \nobreak\hspace{0pt}0.004 [0.001, 0.011] & \nobreak\hspace{0pt}0.003 [0.000, 0.009] & \nobreak\hspace{0pt}0.010 [0.006, 0.014] & \nobreak\hspace{0pt}0.014 [0.014, 0.016] & \nobreak\hspace{0pt}0.013 [0.012, 0.015] \\
\nobreak\hspace{0pt}normal-z-line & \nobreak\hspace{0pt}0.029 [0.022, 0.047] & \nobreak\hspace{0pt}0.023 [0.014, 0.043] & \nobreak\hspace{0pt}0.027 [0.019, 0.047] & \nobreak\hspace{0pt}0.031 [0.016, 0.048] & \nobreak\hspace{0pt}0.016 [0.015, 0.035] & \nobreak\hspace{0pt}0.017 [0.014, 0.023] & \nobreak\hspace{0pt}0.010 [0.008, 0.016] \\
\nobreak\hspace{0pt}polyps & \nobreak\hspace{0pt}0.009 [0.006, 0.021] & \nobreak\hspace{0pt}0.017 [0.009, 0.032] & \nobreak\hspace{0pt}0.012 [0.005, 0.024] & \nobreak\hspace{0pt}0.015 [0.006, 0.028] & \nobreak\hspace{0pt}0.007 [0.004, 0.018] & \nobreak\hspace{0pt}0.011 [0.011, 0.016] & \nobreak\hspace{0pt}0.012 [0.010, 0.016] \\
\nobreak\hspace{0pt}ulcerative-colitis & \nobreak\hspace{0pt}0.012 [0.008, 0.017] & \nobreak\hspace{0pt}0.015 [0.010, 0.027] & \nobreak\hspace{0pt}0.007 [0.003, 0.014] & \nobreak\hspace{0pt}0.010 [0.004, 0.022] & \nobreak\hspace{0pt}0.010 [0.006, 0.015] & \nobreak\hspace{0pt}0.013 [0.011, 0.017] & \nobreak\hspace{0pt}0.016 [0.013, 0.019] \\
\bottomrule
\end{tabular}
\end{table}

\begin{figure}[!htbp]
\centering
\includegraphics[width=1.0\linewidth,height=0.8\textheight,keepaspectratio]{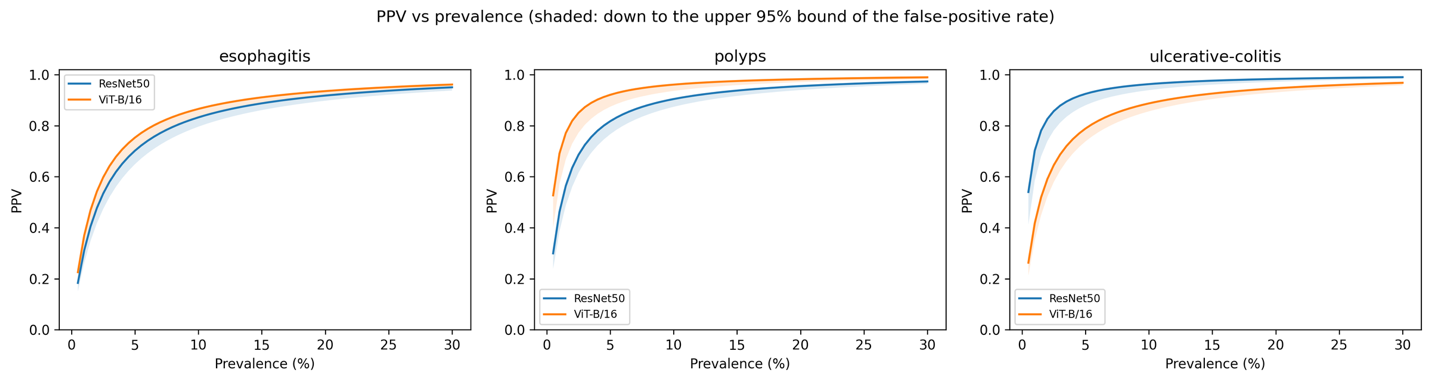}
\caption{Positive predictive value as a function of prevalence for esophagitis, polyps and ulcerative colitis, computed from cross-validated sensitivity and false-positive rate (shaded: down to the upper 95\% bound of the false-positive rate).}
\label{fig:14}
\end{figure}

\textbf{Predictive values under prevalence shift: }Kvasir is balanced, whereas pathological findings are much rarer in routine endoscopy. Using cross-validated sensitivities and false-positive rates (Figure~\ref{fig:14}; Table~\ref{tab:12}), the positive predictive value for esophagitis would fall from 0.87--0.89 at the 12.5\% prevalence of the dataset to 0.70 (ResNet50) and 0.75 (ViT) at 5\% prevalence and to 0.48 and 0.54 at 2\%. For polyps, ViT's lower false-positive rate (0.4\% vs. 1.1\%) gives a higher PPV at 2\% prevalence (0.82 vs. 0.63), whereas for ulcerative colitis the advantage is reversed (0.83 for ResNet50 vs. 0.59 for ViT). Class-level differences that are invisible in balanced accuracy can therefore translate into materially different predictive values at low prevalence. Re-weighting the out-of-fold predictions to an illustrative screening-like prior (75\% normal landmarks; Table~\ref{tab:13}) left calibration low for both models (ECE 0.025 and 0.026), and prior-shift correction of the softmax outputs improved accuracy, Brier score and NLL for both (for example, Brier score 0.105 $\rightarrow$ 0.094 for ResNet50 and 0.102 $\rightarrow$ 0.090 for ViT).

\begin{table}[!htbp]
\centering
\scriptsize
\setlength{\tabcolsep}{4pt}
\renewcommand{\arraystretch}{1.15}
\caption{Positive predictive values (PPV) at 1--20\% prevalence, computed from cross-validated sensitivity and false-positive rate (FPR); ``precision at 12.5\%'' is the observed precision at the dataset prevalence.}
\label{tab:12}
\begin{tabular}{>{\raggedright\arraybackslash}p{0.088\linewidth}>{\raggedright\arraybackslash}p{0.112\linewidth}>{\raggedright\arraybackslash}p{0.074\linewidth}>{\raggedright\arraybackslash}p{0.056\linewidth}>{\raggedright\arraybackslash}p{0.078\linewidth}>{\raggedright\arraybackslash}p{0.074\linewidth}>{\raggedright\arraybackslash}p{0.074\linewidth}>{\raggedright\arraybackslash}p{0.074\linewidth}>{\raggedright\arraybackslash}p{0.074\linewidth}>{\raggedright\arraybackslash}p{0.074\linewidth}}
\toprule
\nobreak\hspace{0pt}\textbf{Model} & \nobreak\hspace{0pt}\textbf{Class} & \nobreak\hspace{0pt}\textbf{Sensitivity} & \nobreak\hspace{0pt}\textbf{FPR} & \nobreak\hspace{0pt}\textbf{Precision at 12.5\%} & \nobreak\hspace{0pt}\textbf{PPV 1\%} & \nobreak\hspace{0pt}\textbf{PPV 2\%} & \nobreak\hspace{0pt}\textbf{PPV 5\%} & \nobreak\hspace{0pt}\textbf{PPV 10\%} & \nobreak\hspace{0pt}\textbf{PPV 20\%} \\
\midrule
\nobreak\hspace{0pt}ResNet50 & \nobreak\hspace{0pt}esophagitis & \nobreak\hspace{0pt}0.818 & \nobreak\hspace{0pt}0.018 & \nobreak\hspace{0pt}0.865 & \nobreak\hspace{0pt}0.311 & \nobreak\hspace{0pt}0.477 & \nobreak\hspace{0pt}0.702 & \nobreak\hspace{0pt}0.833 & \nobreak\hspace{0pt}0.918 \\
\nobreak\hspace{0pt}ResNet50 & \nobreak\hspace{0pt}polyps & \nobreak\hspace{0pt}0.948 & \nobreak\hspace{0pt}0.011 & \nobreak\hspace{0pt}0.924 & \nobreak\hspace{0pt}0.462 & \nobreak\hspace{0pt}0.635 & \nobreak\hspace{0pt}0.817 & \nobreak\hspace{0pt}0.904 & \nobreak\hspace{0pt}0.955 \\
\nobreak\hspace{0pt}ResNet50 & \nobreak\hspace{0pt}ulcerative-colitis & \nobreak\hspace{0pt}0.934 & \nobreak\hspace{0pt}0.004 & \nobreak\hspace{0pt}0.971 & \nobreak\hspace{0pt}0.702 & \nobreak\hspace{0pt}0.827 & \nobreak\hspace{0pt}0.925 & \nobreak\hspace{0pt}0.963 & \nobreak\hspace{0pt}0.983 \\
\nobreak\hspace{0pt}ViT-B/16 & \nobreak\hspace{0pt}esophagitis & \nobreak\hspace{0pt}0.812 & \nobreak\hspace{0pt}0.014 & \nobreak\hspace{0pt}0.892 & \nobreak\hspace{0pt}0.369 & \nobreak\hspace{0pt}0.542 & \nobreak\hspace{0pt}0.753 & \nobreak\hspace{0pt}0.866 & \nobreak\hspace{0pt}0.935 \\
\nobreak\hspace{0pt}ViT-B/16 & \nobreak\hspace{0pt}polyps & \nobreak\hspace{0pt}0.886 & \nobreak\hspace{0pt}0.004 & \nobreak\hspace{0pt}0.969 & \nobreak\hspace{0pt}0.691 & \nobreak\hspace{0pt}0.819 & \nobreak\hspace{0pt}0.921 & \nobreak\hspace{0pt}0.961 & \nobreak\hspace{0pt}0.982 \\
\nobreak\hspace{0pt}ViT-B/16 & \nobreak\hspace{0pt}ulcerative-colitis & \nobreak\hspace{0pt}0.972 & \nobreak\hspace{0pt}0.014 & \nobreak\hspace{0pt}0.910 & \nobreak\hspace{0pt}0.417 & \nobreak\hspace{0pt}0.591 & \nobreak\hspace{0pt}0.789 & \nobreak\hspace{0pt}0.887 & \nobreak\hspace{0pt}0.947 \\
\bottomrule
\end{tabular}
\end{table}

\begin{table}[!htbp]
\centering
\small
\setlength{\tabcolsep}{4pt}
\renewcommand{\arraystretch}{1.15}
\caption{Calibration of the cross-validated predictions re-weighted to an illustrative screening-like class prior (normal landmarks 25\% each; esophagitis and polyps 8\% each; ulcerative colitis and the two dyed classes 3\% each), without and with prior-shift correction of the softmax outputs; 95\% bootstrap confidence intervals.}
\label{tab:13}
\begin{tabular}{>{\raggedright\arraybackslash}p{0.212\linewidth}>{\raggedright\arraybackslash}p{0.173\linewidth}>{\raggedright\arraybackslash}p{0.173\linewidth}>{\raggedright\arraybackslash}p{0.173\linewidth}>{\raggedright\arraybackslash}p{0.173\linewidth}}
\toprule
\nobreak\hspace{0pt}\textbf{Metric} & \nobreak\hspace{0pt}\textbf{ResNet50 unadjusted} & \nobreak\hspace{0pt}\textbf{ResNet50 prior-shift corrected} & \nobreak\hspace{0pt}\textbf{ViT-B/16 unadjusted} & \nobreak\hspace{0pt}\textbf{ViT-B/16 prior-shift corrected} \\
\midrule
\nobreak\hspace{0pt}Accuracy & \nobreak\hspace{0pt}0.930 [0.922, 0.938] & \nobreak\hspace{0pt}0.944 [0.937, 0.951] & \nobreak\hspace{0pt}0.934 [0.927, 0.942] & \nobreak\hspace{0pt}0.948 [0.940, 0.954] \\
\nobreak\hspace{0pt}ECE, 15 equal-width bins & \nobreak\hspace{0pt}0.025 [0.020, 0.033] & \nobreak\hspace{0pt}0.021 [0.015, 0.028] & \nobreak\hspace{0pt}0.026 [0.023, 0.033] & \nobreak\hspace{0pt}0.023 [0.019, 0.029] \\
\nobreak\hspace{0pt}ECE, 10 equal-width bins & \nobreak\hspace{0pt}0.021 [0.017, 0.029] & \nobreak\hspace{0pt}0.020 [0.014, 0.027] & \nobreak\hspace{0pt}0.027 [0.022, 0.034] & \nobreak\hspace{0pt}0.021 [0.016, 0.028] \\
\nobreak\hspace{0pt}ECE, 15 equal-mass bins & \nobreak\hspace{0pt}0.025 [0.020, 0.032] & \nobreak\hspace{0pt}0.022 [0.018, 0.030] & \nobreak\hspace{0pt}0.024 [0.020, 0.031] & \nobreak\hspace{0pt}0.022 [0.019, 0.030] \\
\nobreak\hspace{0pt}Class-wise ECE & \nobreak\hspace{0pt}0.016 [0.015, 0.018] & \nobreak\hspace{0pt}0.012 [0.012, 0.015] & \nobreak\hspace{0pt}0.016 [0.015, 0.018] & \nobreak\hspace{0pt}0.010 [0.010, 0.013] \\
\nobreak\hspace{0pt}Brier score & \nobreak\hspace{0pt}0.105 [0.095, 0.117] & \nobreak\hspace{0pt}0.094 [0.084, 0.105] & \nobreak\hspace{0pt}0.102 [0.093, 0.113] & \nobreak\hspace{0pt}0.090 [0.080, 0.100] \\
\nobreak\hspace{0pt}NLL & \nobreak\hspace{0pt}0.231 [0.211, 0.252] & \nobreak\hspace{0pt}0.213 [0.192, 0.234] & \nobreak\hspace{0pt}0.224 [0.206, 0.243] & \nobreak\hspace{0pt}0.202 [0.184, 0.222] \\
\bottomrule
\end{tabular}
\end{table}

\subsection{Interpretability and Reliance on an Acquisition Artefact}

\begin{figure}[!htbp]
\centering
\includegraphics[width=1.0\linewidth,height=0.4\textheight,keepaspectratio]{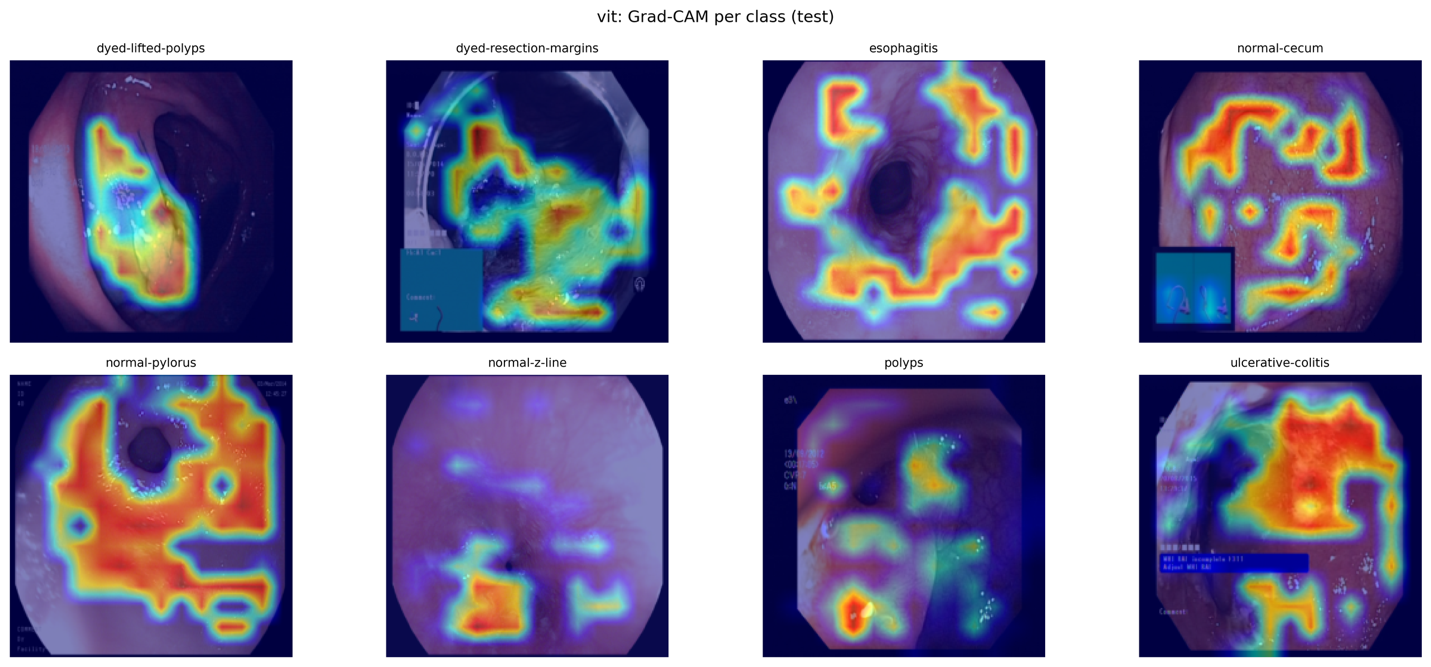}
\\[4pt]
\includegraphics[width=1.0\linewidth,height=0.4\textheight,keepaspectratio]{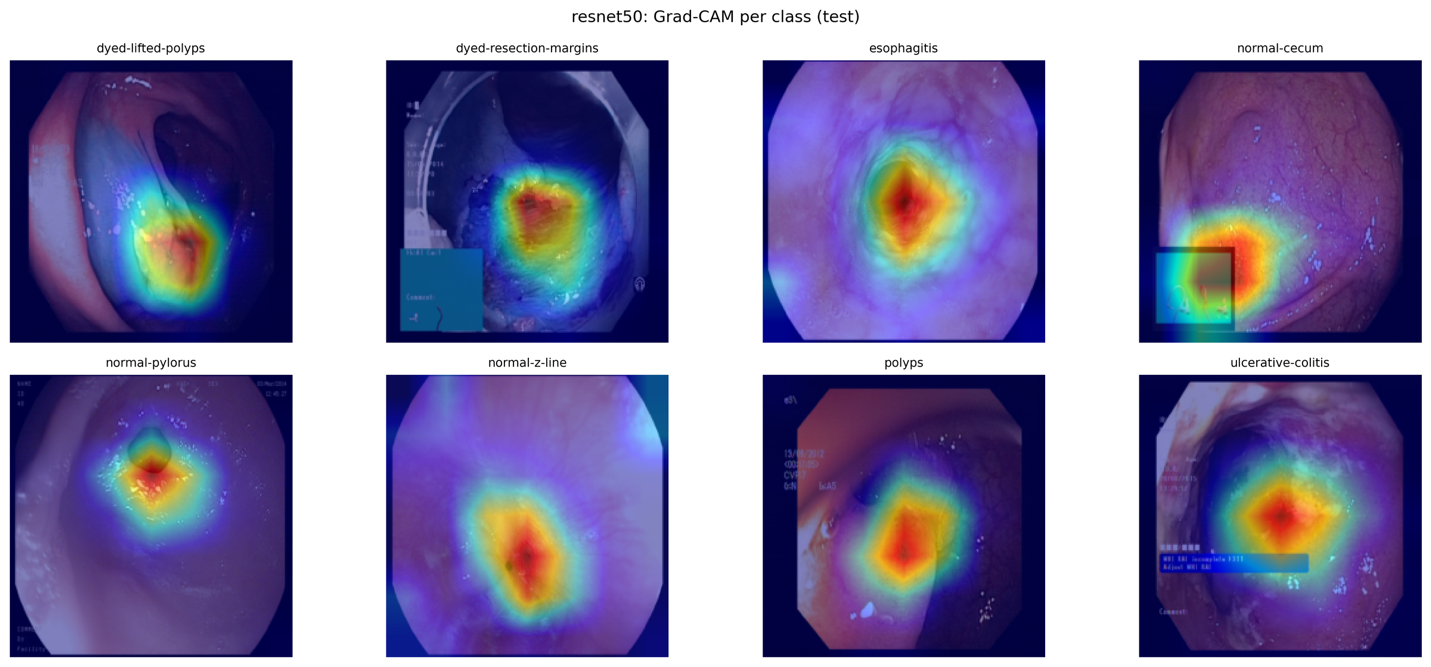}
\caption{Grad-CAM heatmaps of ViT-B/16 (top) and ResNet50 (bottom) for the first test image of each class. Note the ResNet50 activation on the ScopeGuide inset in the normal-cecum example.}
\label{fig:15}
\end{figure}

\begin{figure}[!htbp]
\centering
\includegraphics[width=1.0\linewidth,height=0.4\textheight,keepaspectratio]{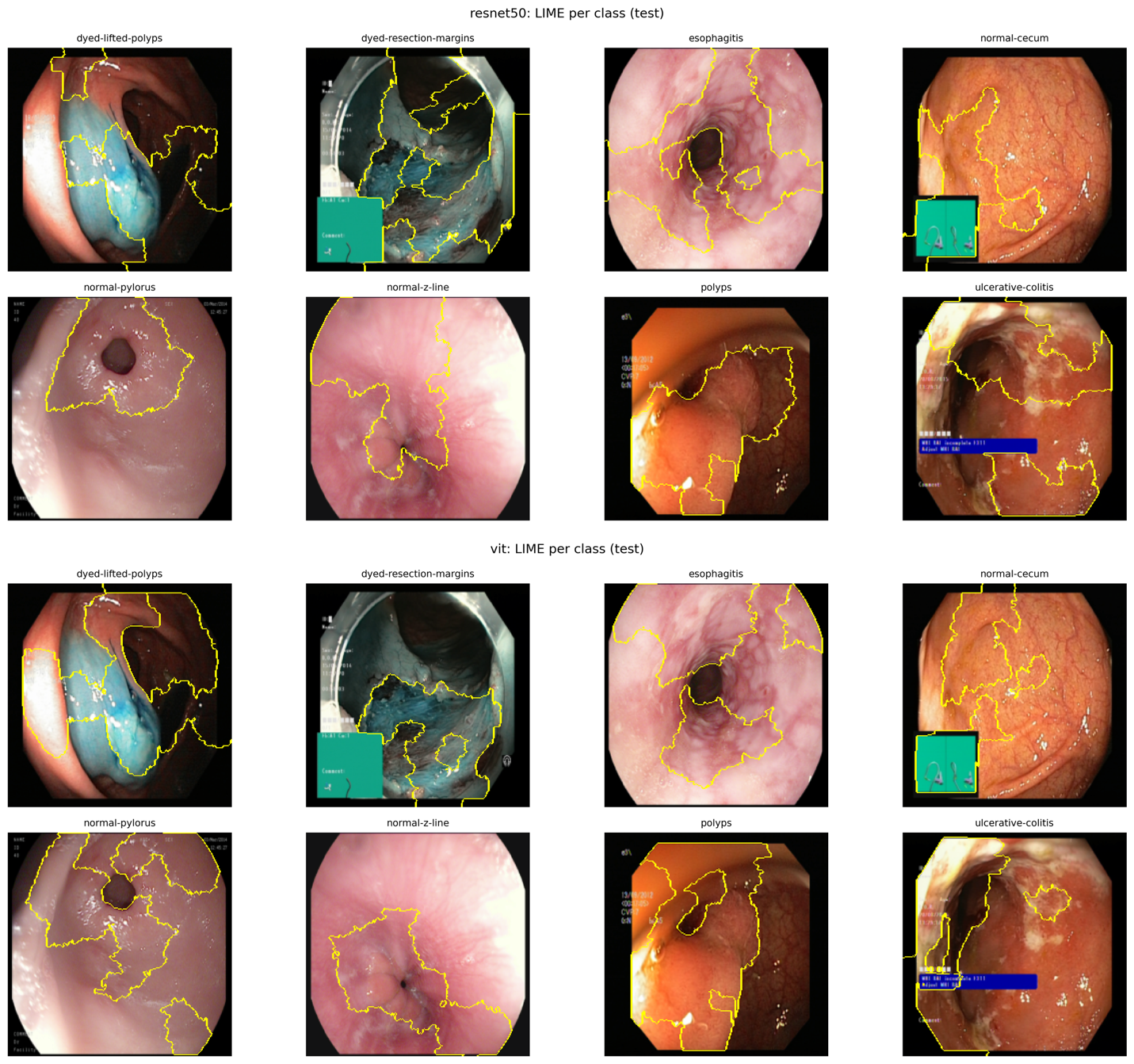}
\\[4pt]
\includegraphics[width=1.0\linewidth,height=0.4\textheight,keepaspectratio]{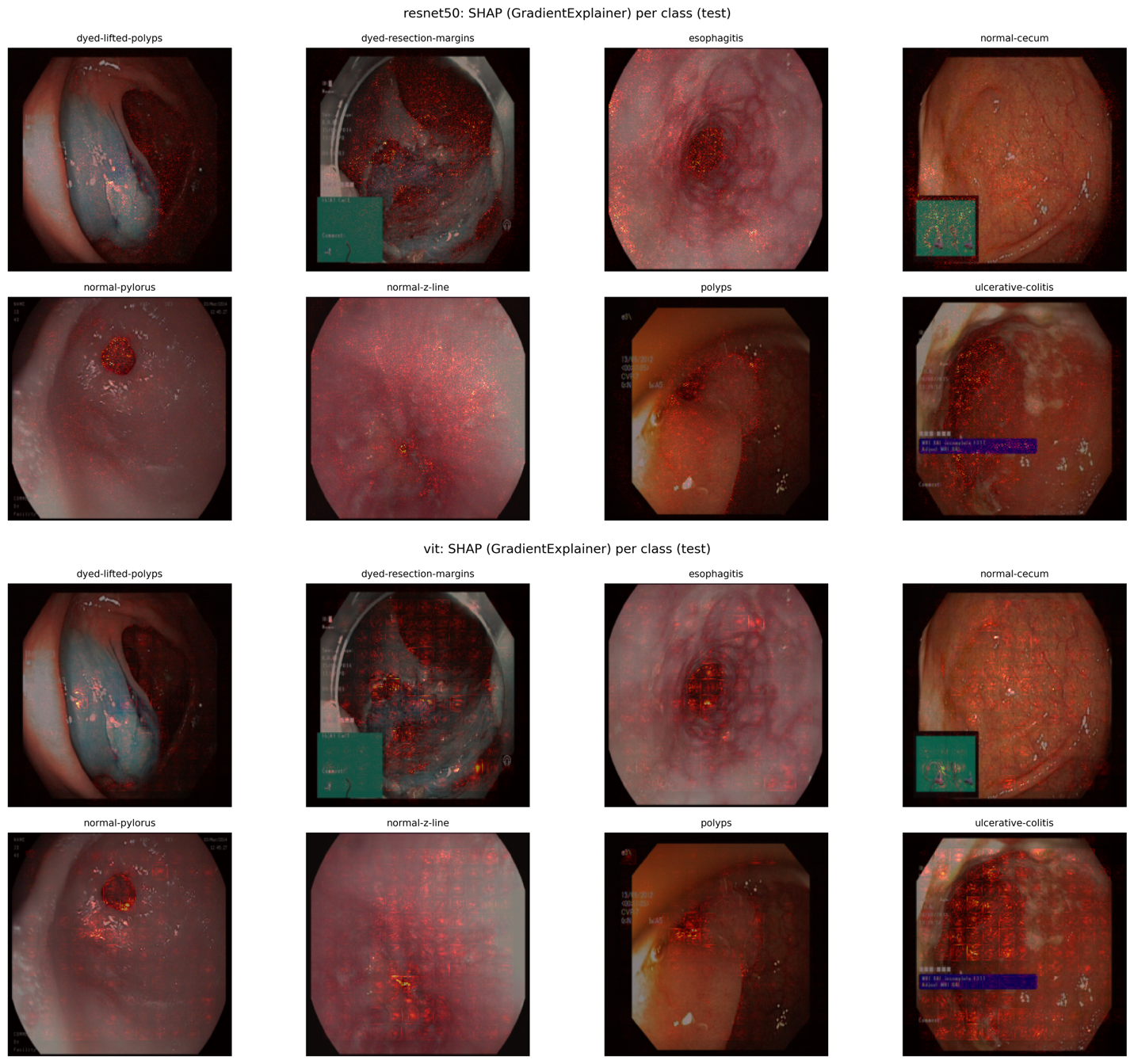}
\caption{LIME attributions (first image; ResNet50 above ViT-B/16) and SHAP GradientExplainer attributions (second image; ResNet50 above ViT-B/16) for the first test image of each class.}
\label{fig:16}
\end{figure}

\begin{figure}[!htbp]
\centering
\includegraphics[width=1.0\linewidth,height=0.8\textheight,keepaspectratio]{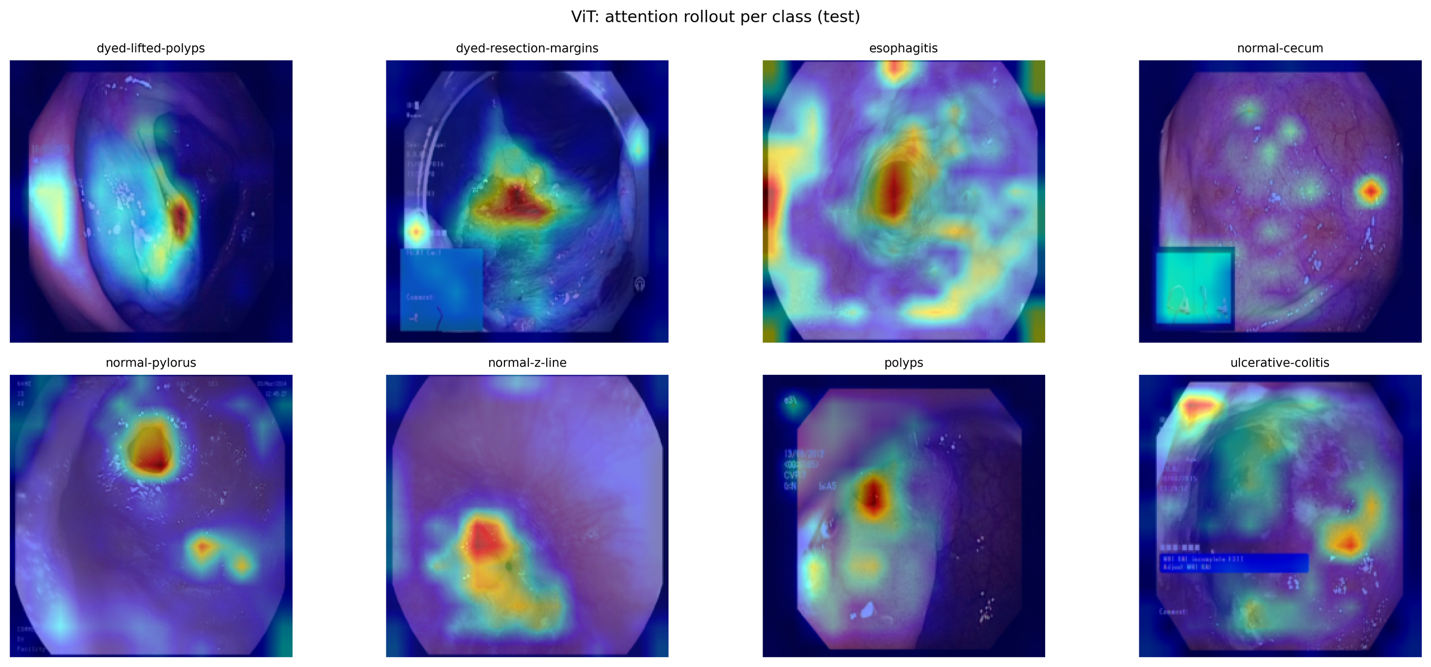}
\caption{ViT-B/16 attention-rollout maps for the first test image of each class.}
\label{fig:17}
\end{figure}

Grad-CAM heatmaps for representative test images are shown in Figure~\ref{fig:15}, LIME and SHAP attributions in Figure~\ref{fig:16}, and ViT attention rollout in Figure~\ref{fig:17}. For most classes, the maps of both models concentrate on the central mucosal field. The figures also reveal a potential shortcut: in the normal-cecum example, ResNet50's Grad-CAM concentrates on the green ScopeGuide inset in the lower-left corner rather than on the mucosa.

\begin{figure}[!htbp]
\centering
\includegraphics[width=1.0\linewidth,height=0.8\textheight,keepaspectratio]{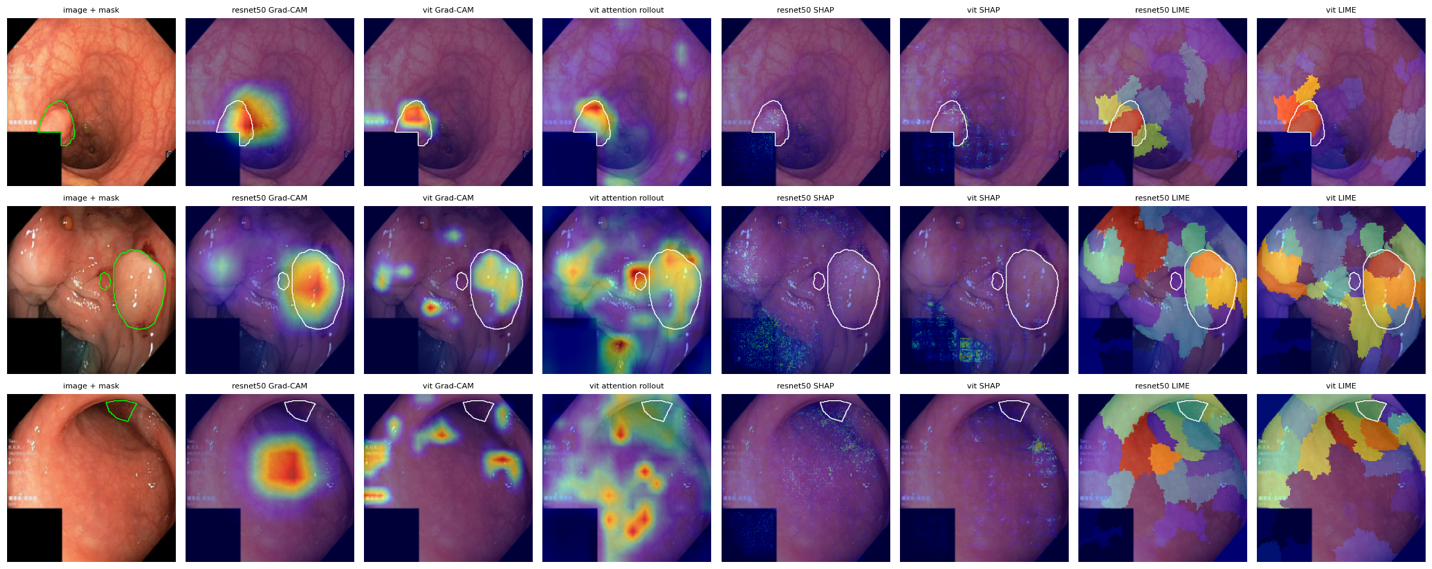}
\caption{Lesion-level localization on unseen Kvasir-SEG polyp images. First column: image with the expert polyp mask (green contour); remaining columns: Grad-CAM (ResNet50, ViT-B/16), ViT-B/16 attention rollout, SHAP (ResNet50, ViT-B/16) and LIME (ResNet50, ViT-B/16) maps for the class polyps, with the mask outlined in white. The black square is the ScopeGuide region, which is removed in Kvasir-SEG.}
\label{fig:18}
\end{figure}

\textbf{Lesion-level localization:} On 500 unseen Kvasir-SEG polyp images (Table~\ref{tab:14}, Figure~\ref{fig:18}), both models concentrated attribution on the lesion well above chance. With Grad-CAM, the energy-in-lesion fraction was 0.553 [0.533, 0.573] for ResNet50 and 0.575 [0.552, 0.599] for ViT, against a chance level of 0.154 (the mean lesion area), and the attribution maximum fell inside the lesion in 86.8\% and 72.8\% of images. ResNet50's peak was therefore more often on the polyp (102 vs. 32 discordant images; exact p \textless{} 0.001), whereas ViT placed slightly more of its total attribution inside it (Wilcoxon p = 0.002). LIME favoured ResNet50 (energy-in-lesion 0.431 vs. 0.325, p \textless{} 0.001; n = 50), SHAP was the least localized method and did not differ between the models (0.270 vs. 0.252, p = 0.10), and ViT attention rollout was less focused than Grad-CAM (0.325). Both models also classified these unseen images correctly as polyps (97.8\% and 95.8\%), although the ScopeGuide region is blacked out in Kvasir-SEG. These findings agree with earlier work: Grad-CAM has been used to make polyp segmentation on Kvasir images transparent \cite{r50}, and in diabetic foot ulcer segmentation a quantitative comparison against expert masks found Grad-CAM more lesion-localized than attention rollout, whereas attention rollout was more faithful to the model's computation \cite{r51}. ViT showed the same localization ordering here (energy-in-lesion 0.575 for Grad-CAM vs. 0.325 for attention rollout).

\begin{table}[!htbp]
\centering
\footnotesize
\setlength{\tabcolsep}{4pt}
\renewcommand{\arraystretch}{1.15}
\caption{Lesion-level localization of the class polyps on unseen Kvasir-SEG images (mean [95\% bootstrap CI]). The chance level of energy-in-lesion equals the lesion's area fraction.}
\label{tab:14}
\begin{tabular}{>{\raggedright\arraybackslash}p{0.212\linewidth}>{\raggedright\arraybackslash}p{0.045\linewidth}>{\raggedright\arraybackslash}p{0.157\linewidth}>{\raggedright\arraybackslash}p{0.157\linewidth}>{\raggedright\arraybackslash}p{0.157\linewidth}>{\raggedright\arraybackslash}p{0.157\linewidth}}
\toprule
\nobreak\hspace{0pt}\textbf{Method} & \nobreak\hspace{0pt}\textbf{n} & \nobreak\hspace{0pt}\textbf{Pointing-game hit rate} & \nobreak\hspace{0pt}\textbf{Energy-in-lesion} & \nobreak\hspace{0pt}\textbf{Chance level (lesion area)} & \nobreak\hspace{0pt}\textbf{IoU (top-20\% map)} \\
\midrule
\nobreak\hspace{0pt}ResNet50 Grad-CAM & \nobreak\hspace{0pt}500 & \nobreak\hspace{0pt}0.868 [0.838, 0.896] & \nobreak\hspace{0pt}0.553 [0.533, 0.573] & \nobreak\hspace{0pt}0.154 [0.143, 0.165] & \nobreak\hspace{0pt}0.396 [0.380, 0.412] \\
\nobreak\hspace{0pt}ViT-B/16 Grad-CAM & \nobreak\hspace{0pt}500 & \nobreak\hspace{0pt}0.728 [0.688, 0.766] & \nobreak\hspace{0pt}0.575 [0.552, 0.599] & \nobreak\hspace{0pt}0.154 [0.143, 0.165] & \nobreak\hspace{0pt}0.350 [0.333, 0.367] \\
\nobreak\hspace{0pt}ViT-B/16 attention rollout & \nobreak\hspace{0pt}500 & \nobreak\hspace{0pt}0.624 [0.582, 0.666] & \nobreak\hspace{0pt}0.325 [0.309, 0.340] & \nobreak\hspace{0pt}0.154 [0.142, 0.165] & \nobreak\hspace{0pt}0.337 [0.323, 0.350] \\
\nobreak\hspace{0pt}ResNet50 SHAP & \nobreak\hspace{0pt}50 & \nobreak\hspace{0pt}0.500 [0.360, 0.640] & \nobreak\hspace{0pt}0.270 [0.221, 0.323] & \nobreak\hspace{0pt}0.161 [0.122, 0.204] & \nobreak\hspace{0pt}0.207 [0.179, 0.235] \\
\nobreak\hspace{0pt}ViT-B/16 SHAP & \nobreak\hspace{0pt}50 & \nobreak\hspace{0pt}0.500 [0.360, 0.640] & \nobreak\hspace{0pt}0.252 [0.206, 0.301] & \nobreak\hspace{0pt}0.161 [0.124, 0.203] & \nobreak\hspace{0pt}0.173 [0.150, 0.197] \\
\nobreak\hspace{0pt}ResNet50 LIME & \nobreak\hspace{0pt}50 & \nobreak\hspace{0pt}0.600 [0.460, 0.740] & \nobreak\hspace{0pt}0.431 [0.365, 0.495] & \nobreak\hspace{0pt}0.161 [0.123, 0.203] & \nobreak\hspace{0pt}0.301 [0.255, 0.349] \\
\nobreak\hspace{0pt}ViT-B/16 LIME & \nobreak\hspace{0pt}50 & \nobreak\hspace{0pt}0.580 [0.440, 0.720] & \nobreak\hspace{0pt}0.325 [0.276, 0.376] & \nobreak\hspace{0pt}0.161 [0.125, 0.204] & \nobreak\hspace{0pt}0.244 [0.202, 0.285] \\
\bottomrule
\end{tabular}
\end{table}

\textbf{Reliance on the ScopeGuide inset.} The inset was present in 149 of the 400 test images (normal-cecum 49, dyed-resection-margins 38, dyed-lifted-polyps 29, polyps 26, ulcerative-colitis 7). Blacking it out (Table~\ref{tab:15}) did not significantly affect ResNet50 (95.3\% $\rightarrow$ 94.0\%; 4 images lost, 2 gained; McNemar p = 0.69), reduced ViT's accuracy by 4.0 points (93.3\% $\rightarrow$ 89.3\%; p = 0.15), and reduced MobileViT-S's accuracy by 9.4 points (96.6\% $\rightarrow$ 87.2\%; 14 images lost, none gained; p \textless{} 0.001). MobileViT-S's advantage on the primary split is therefore at least partly explained by reliance on an acquisition artefact rather than on pathology, and ResNet50 was the least sensitive of the three models. Because the inset occurs only in lower-GI and procedural classes, any model can use it to separate these from the upper-GI classes; the occlusion test is a simple, reproducible check that we recommend for any Kvasir benchmark.

\begin{table}[!htbp]
\centering
\footnotesize
\setlength{\tabcolsep}{4pt}
\renewcommand{\arraystretch}{1.15}
\caption{Shortcut test: accuracy on the 149 test images containing the ScopeGuide inset, before and after blacking out the inset.}
\label{tab:15}
\begin{tabular}{>{\raggedright\arraybackslash}p{0.158\linewidth}>{\raggedright\arraybackslash}p{0.056\linewidth}>{\raggedright\arraybackslash}p{0.130\linewidth}>{\raggedright\arraybackslash}p{0.141\linewidth}>{\raggedright\arraybackslash}p{0.120\linewidth}>{\raggedright\arraybackslash}p{0.120\linewidth}>{\raggedright\arraybackslash}p{0.141\linewidth}}
\toprule
\nobreak\hspace{0pt}\textbf{Model} & \nobreak\hspace{0pt}\textbf{n} & \nobreak\hspace{0pt}\textbf{Accuracy, original (\%)} & \nobreak\hspace{0pt}\textbf{Accuracy, inset occluded (\%)} & \nobreak\hspace{0pt}\textbf{Prediction flips (\%)} & \nobreak\hspace{0pt}\textbf{Lost / gained} & \nobreak\hspace{0pt}\textbf{McNemar exact p} \\
\midrule
\nobreak\hspace{0pt}ResNet50 & \nobreak\hspace{0pt}149 & \nobreak\hspace{0pt}95.30 & \nobreak\hspace{0pt}93.96 & \nobreak\hspace{0pt}4.7 & \nobreak\hspace{0pt}4 / 2 & \nobreak\hspace{0pt}0.69 \\
\nobreak\hspace{0pt}ViT-B/16 & \nobreak\hspace{0pt}149 & \nobreak\hspace{0pt}93.29 & \nobreak\hspace{0pt}89.26 & \nobreak\hspace{0pt}8.1 & \nobreak\hspace{0pt}9 / 3 & \nobreak\hspace{0pt}0.15 \\
\nobreak\hspace{0pt}MobileViT-S & \nobreak\hspace{0pt}149 & \nobreak\hspace{0pt}96.64 & \nobreak\hspace{0pt}87.25 & \nobreak\hspace{0pt}10.7 & \nobreak\hspace{0pt}14 / 0 & \nobreak\hspace{0pt}\textless{} 0.001 \\
\bottomrule
\end{tabular}
\end{table}

\textbf{Attribution on the inset.} In 40 of the test images containing the inset, ResNet50's maps placed slightly more attribution on the inset than its area fraction (0.183) would predict: 1.19 times chance for Grad-CAM, 1.32 for LIME and 1.36 for SHAP (Table~\ref{tab:16}). ViT's Grad-CAM and SHAP maps placed less than chance on it (0.73 and 0.82 times), whereas its attention rollout placed more (1.31 times). Attribution mass alone therefore does not show which model depends on the artefact: ResNet50 attends to the inset to some extent, yet its predictions were the least affected by removing it (Table~\ref{tab:15}).

\begin{table}[!htbp]
\centering
\small
\setlength{\tabcolsep}{4pt}
\renewcommand{\arraystretch}{1.15}
\caption{Share of attribution mass falling on the ScopeGuide inset (mean over 40 test images containing the inset), compared with the inset's area fraction, the level expected for a map that ignores the inset.}
\label{tab:16}
\begin{tabular}{>{\raggedright\arraybackslash}p{0.308\linewidth}>{\raggedright\arraybackslash}p{0.205\linewidth}>{\raggedright\arraybackslash}p{0.205\linewidth}>{\raggedright\arraybackslash}p{0.205\linewidth}}
\toprule
\nobreak\hspace{0pt}\textbf{Method} & \nobreak\hspace{0pt}\textbf{Attribution mass on the inset} & \nobreak\hspace{0pt}\textbf{Inset area fraction (chance level)} & \nobreak\hspace{0pt}\textbf{Ratio to chance} \\
\midrule
\nobreak\hspace{0pt}ResNet50 Grad-CAM & \nobreak\hspace{0pt}0.217 & \nobreak\hspace{0pt}0.183 & \nobreak\hspace{0pt}1.19 \\
\nobreak\hspace{0pt}ResNet50 LIME & \nobreak\hspace{0pt}0.242 & \nobreak\hspace{0pt}0.183 & \nobreak\hspace{0pt}1.32 \\
\nobreak\hspace{0pt}ResNet50 SHAP & \nobreak\hspace{0pt}0.248 & \nobreak\hspace{0pt}0.183 & \nobreak\hspace{0pt}1.36 \\
\nobreak\hspace{0pt}ViT-B/16 Grad-CAM & \nobreak\hspace{0pt}0.133 & \nobreak\hspace{0pt}0.183 & \nobreak\hspace{0pt}0.73 \\
\nobreak\hspace{0pt}ViT-B/16 LIME & \nobreak\hspace{0pt}0.203 & \nobreak\hspace{0pt}0.183 & \nobreak\hspace{0pt}1.11 \\
\nobreak\hspace{0pt}ViT-B/16 SHAP & \nobreak\hspace{0pt}0.151 & \nobreak\hspace{0pt}0.183 & \nobreak\hspace{0pt}0.82 \\
\nobreak\hspace{0pt}ViT-B/16 attention rollout & \nobreak\hspace{0pt}0.239 & \nobreak\hspace{0pt}0.183 & \nobreak\hspace{0pt}1.31 \\
\bottomrule
\end{tabular}
\end{table}

\subsection{External Evaluation on GastroVision}

\begin{figure}[!htbp]
\centering
\includegraphics[width=1.0\linewidth,height=0.8\textheight,keepaspectratio]{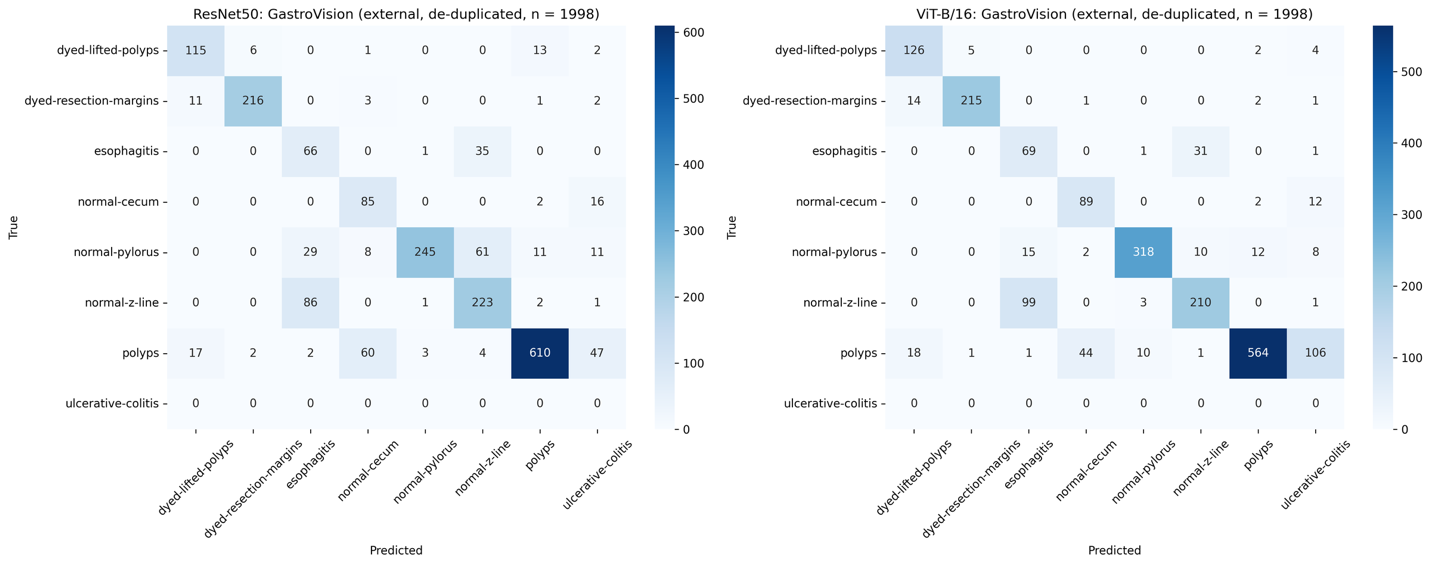}
\caption{Confusion matrices of ResNet50 (left) and ViT-B/16 (right) on the de-duplicated GastroVision subset (n = 1,998). GastroVision has no ulcerative-colitis images, so that row is empty and every prediction of ulcerative colitis is an error.}
\label{fig:19}
\end{figure}

On the 1,998 de-duplicated GastroVision images (Table~\ref{tab:17}), accuracy fell to 78.08\% for ResNet50, 79.63\% for ViT and 78.33\% for MobileViT-S, 13--17 points below the internal results. ViT was 1.55 points more accurate than ResNet50 (95\% CI $-$3.45 to +0.35; McNemar p = 0.12), and equivalence within $\pm$3 points was narrowly missed (90\% CI $-$3.10 to +0.05; TOST p = 0.067). MobileViT-S was equivalent to ResNet50 (+0.25 points; TOST p = 0.001), so its advantage on the internal split did not transfer, consistent with the shortcut test. Per-class sensitivity varied much more between models than overall accuracy did: ViT was more sensitive for normal-pylorus (0.87 vs. 0.67) and dyed-lifted-polyps (0.92 vs. 0.84), ResNet50 for polyps (0.82 vs. 0.76) and normal-z-line (0.71 vs. 0.67), and esophagitis remained the weakest class for all models (0.57--0.68). The external set is naturally imbalanced (polyps 37\%, esophagitis 5\%), so it also illustrates the prevalence effects discussed in Section 4. Specifically, this severe class imbalance exposes the vulnerabilities of models trained on artificially balanced data; model calibration and class-wise performance degraded sharply on minority classes like esophagitis (sensitivity of 0.57--0.68). This highlights how models can become systematically under-sensitive or miscalibrated when encountering the skewed prior probabilities characteristic of real-world clinical distributions. The confusion matrices (Figure~\ref{fig:19}) show that the external errors follow the same pattern as on Kvasir: esophagitis and normal-z-line were confused in both directions (35 and 86 images for ResNet50; 31 and 99 for ViT). ResNet50 labelled 61 pylorus images as normal-z-line, which explains its low pylorus sensitivity, and mislabelled polyps mainly as normal-cecum (60) and ulcerative colitis (47), whereas ViT labelled 106 polyps as ulcerative colitis, a class that does not occur in GastroVision.

\begin{table}[!htbp]
\centering
\small
\setlength{\tabcolsep}{4pt}
\renewcommand{\arraystretch}{1.15}
\caption{External evaluation on GastroVision (seven classes shared with Kvasir; 1,998 images after removing 152 near-duplicates of Kvasir images): per-class sensitivity and overall performance.}
\label{tab:17}
\begin{tabular}{>{\raggedright\arraybackslash}p{0.254\linewidth}>{\raggedright\arraybackslash}p{0.152\linewidth}>{\raggedright\arraybackslash}p{0.141\linewidth}>{\raggedright\arraybackslash}p{0.178\linewidth}>{\raggedright\arraybackslash}p{0.178\linewidth}}
\toprule
\nobreak\hspace{0pt}\textbf{Class} & \nobreak\hspace{0pt}\textbf{n (prevalence)} & \nobreak\hspace{0pt}\textbf{ResNet50} & \nobreak\hspace{0pt}\textbf{ViT-B/16} & \nobreak\hspace{0pt}\textbf{Mobile\-ViT-S} \\
\midrule
\nobreak\hspace{0pt}dyed-lifted-polyps & \nobreak\hspace{0pt}137 (6.9\%) & \nobreak\hspace{0pt}0.839 & \nobreak\hspace{0pt}0.920 & \nobreak\hspace{0pt}0.920 \\
\nobreak\hspace{0pt}dyed-resection-margins & \nobreak\hspace{0pt}233 (11.7\%) & \nobreak\hspace{0pt}0.927 & \nobreak\hspace{0pt}0.923 & \nobreak\hspace{0pt}0.914 \\
\nobreak\hspace{0pt}esophagitis & \nobreak\hspace{0pt}102 (5.1\%) & \nobreak\hspace{0pt}0.647 & \nobreak\hspace{0pt}0.677 & \nobreak\hspace{0pt}0.569 \\
\nobreak\hspace{0pt}normal-cecum & \nobreak\hspace{0pt}103 (5.2\%) & \nobreak\hspace{0pt}0.825 & \nobreak\hspace{0pt}0.864 & \nobreak\hspace{0pt}0.786 \\
\nobreak\hspace{0pt}normal-pylorus & \nobreak\hspace{0pt}365 (18.3\%) & \nobreak\hspace{0pt}0.671 & \nobreak\hspace{0pt}0.871 & \nobreak\hspace{0pt}0.836 \\
\nobreak\hspace{0pt}normal-z-line & \nobreak\hspace{0pt}313 (15.7\%) & \nobreak\hspace{0pt}0.713 & \nobreak\hspace{0pt}0.671 & \nobreak\hspace{0pt}0.744 \\
\nobreak\hspace{0pt}polyps & \nobreak\hspace{0pt}745 (37.3\%) & \nobreak\hspace{0pt}0.819 & \nobreak\hspace{0pt}0.757 & \nobreak\hspace{0pt}0.737 \\
\nobreak\hspace{0pt}Overall accuracy (\%) & \nobreak\hspace{0pt}1,998 & \nobreak\hspace{0pt}78.08 & \nobreak\hspace{0pt}79.63 & \nobreak\hspace{0pt}78.33 \\
\nobreak\hspace{0pt}Macro-averaged recall &  & \nobreak\hspace{0pt}0.777 & \nobreak\hspace{0pt}0.812 & \nobreak\hspace{0pt}0.787 \\
\nobreak\hspace{0pt}$\Delta$ vs. ResNet50 [95\% CI] (pp); McNemar p &  & \nobreak\hspace{0pt}--- & \nobreak\hspace{0pt}+1.55 [$-$0.35, +3.45]; 0.12 & \nobreak\hspace{0pt}+0.25 [$-$1.55, +2.00]; 0.83 \\
\bottomrule
\end{tabular}
\end{table}

\FloatBarrier

\subsection{Discussion}

The central finding is that, under a controlled and reproducible protocol, ResNet50 and ViT-B/16 are equivalent in accuracy for 8-class GI classification on Kvasir: in cross-validation over all 4,000 images the difference was 0.43 points with a 90\% CI of $-$0.33 to +1.18, well inside a $\pm$3-point margin. The single 400-image split, by contrast, could show neither a difference nor equivalence, which illustrates why a non-significant test on one split should not be read as evidence of equivalence. Much of the literature frames the CNN-versus-transformer question in terms of which architecture is more accurate; in this common transfer-learning regime on a moderately sized, curated dataset, the accuracy difference between the two is small.

Once accuracy is set aside as non-discriminating, efficiency decides, and here the architectures differ substantially: ResNet50 delivers equivalent accuracy with 3.6 times fewer parameters and 4.3 times fewer operations, 2.5 times lower CPU latency, 3.8 times higher GPU throughput, and about a quarter of the training time to its best checkpoint. For the resource-constrained settings that motivate this work, where compute, memory, energy and latency are genuine constraints, this makes the CNN the more defensible default relative to ViT-B/16. While our study positions MobileViT-S as the primary edge-optimized baseline to evaluate the transformer paradigm, future evaluations incorporating modern lightweight CNN architectures, such as EfficientNet or MobileNetV3, would further clarify whether the efficiency-accuracy trade-off can be optimized even further without relying on transformer components. ViT remains a reasonable choice where larger and more diverse training data are available, and it was slightly, though not significantly, more accurate on the external data.

Two further results qualify this conclusion. First, robustness to the ScopeGuide inset differed sharply between architectures. MobileViT-S, the most accurate and most CPU-efficient model on the primary split, lost 9.4 points when the inset was occluded and lost its advantage on external data, whereas ResNet50 was not significantly affected. Accuracy on a curated benchmark can therefore reward shortcut learning, and an edge-optimized model should not be selected on benchmark accuracy alone. To mitigate this, future training pipelines must proactively integrate targeted augmentation techniques such as applying random Cutout or explicitly masking the green ScopeGuide box during training to prevent models from exploiting non-pathological features. The interpretability analyses point the same way: both models localized polyps well above chance on unseen images, and attribution mass on the inset did not predict which model depended on it, so attribution maps alone did not separate the architectures, whereas the occlusion test did. Second, every model lost 13--17 points on partially independent external data, an order of magnitude more than the differences between architectures, and predictive values change substantially when prevalence shifts. Architecture choice should therefore be driven by deployment cost and explicit robustness checks, and no model should be deployed without local validation.

Our findings are broadly consistent with the subset of prior comparative studies that report similar accuracies for CNNs and transformers on Kvasir-family data (for example, the high but comparable accuracies reported by Regmi et al. \cite{r27}, Wang et al. \cite{r31}, and Demirbaş et al. \cite{r6}), while adding equivalence testing, cross-validation, a standardized efficiency benchmark, a shortcut test and an external evaluation, which those studies generally do not report. Rather than introducing a new architecture or claiming a new state-of-the-art number, the contribution here is methodological and practical: a reproducible, statistically tested, efficiency-aware benchmark that yields actionable model-selection guidance.

\FloatBarrier

\subsection{Limitations}

This study has several limitations. First, the external evaluation is only partly independent: GastroVision shares a hospital (Bærum) with Kvasir, covers seven of the eight classes, and does not record the centre of each image, so it cannot replace a genuinely independent, ideally prospective, multi-centre validation. Second, Kvasir provides no patient, examination or video identifiers, so a patient-level split was impossible; perceptual-hash grouping limits but cannot eliminate correlated images across folds, and it is deliberately conservative (130 of the 255 candidate pairs crossed class boundaries, indicating that frame layout and the inset also drive hash similarity, although visual inspection confirmed that the top-ranked pairs are identical images). Third, we used the public 4,000-image Kvasir v1 release, which is balanced, so predictive values and calibration do not transfer directly to screening populations (Section 4.5). Fourth, the comparison uses standard matched training configurations rather than exhaustively tuned ones, MobileViT-S was evaluated only on the primary split, and the equivalence finding may shift under other training regimes or larger datasets. Fifth, the interpretability analyses remain partly qualitative: lesion-level annotations exist only for polyps, and the attribution maps have not yet been rated by endoscopists. A blinded rating package (randomised, model- and method-anonymised Grad-CAM, SHAP, LIME and attention-rollout maps) has been prepared for evaluation by gastroenterologists. Finally, efficiency was measured on one GPU and one cloud CPU and should be re-measured on the intended deployment device.

\section{Conclusion}

This study presented a controlled, reproducible benchmark of ResNet50 and ViT-B/16 for 8-class GI classification on the public Kvasir dataset (4,000 images). On a fixed 400-image test split the models did not differ significantly (93.75\% vs. 92.25\%; McNemar p = 0.21), and stratified group 5-fold cross-validation over all images showed them to be equivalent within $\pm$3 percentage points (92.12\% vs. 91.70\%; 90\% CI for the difference $-$0.33 to +1.18; TOST p \textless{} 0.001), with similar calibration. They differed substantially in cost: ResNet50 used 3.6 times fewer parameters and 4.3 times fewer operations, ran 2.5 times faster on a CPU with 3.8 times higher GPU throughput, and reached its best checkpoint about four times sooner. An edge-oriented transformer, MobileViT-S, was most accurate on the test split but depended strongly on an acquisition artefact, and all models lost 13--17 points on external data. The Grad-CAM maps of both CNN and transformer localized polyps on unseen images at about 3.6 times the chance level. Accuracy differences between architectures were thus small compared with the effects of shortcut learning and domain shift, while efficiency differences were large. ResNet50 is a defensible default for resource-constrained GI screening, provided that models are checked for artefact reliance and validated locally.

Future work should prioritise: (1) prospective, multi-centre external validation with patient-level data and realistic prevalence; (2) completion of the blinded expert rating of attribution maps and lesion-level annotations for all pathological classes; (3) training strategies that proactively remove artefact shortcuts, such as applying random Cutout or explicitly masking the ScopeGuide inset during training pipelines; (4) evaluation of edge-optimized and quantized models on target hardware; and (5) prior-shift recalibration for deployment at local prevalence.

\section*{Acknowledgements}

The authors would like to express their gratitude to the creators and maintainers of the Kvasir dataset for providing a high-quality, publicly available benchmark that has facilitated significant advancements in gastrointestinal disease detection research. We also extend our appreciation to colleagues and collaborators whose constructive feedback and insights supported the development of this work.

During the revision, the authors used Claude (Anthropic) to help draft revised text, the response to reviewers, and analysis code. The authors reviewed, verified and edited all AI-assisted content, ran all experiments, and take full responsibility for the manuscript.

\section*{Data Availability}

The dataset employed in this study, Kvasir, is openly accessible to the research community. The release used here contains 4,000 annotated endoscopic images across eight clinically relevant classes and can be freely obtained through its official publication \cite{r17} and the hosting platform: \url{https://datasets.simula.no/kvasir/}; the copy analysed in this study was obtained from \url{https://www.kaggle.com/datasets/meetnagadia/kvasir-dataset}. Kvasir-SEG and GastroVision are publicly available from their authors \cite{r19,r48}. The analysis notebook, fixed data split, cross-validation fold assignments and result tables are available from the corresponding author on request.

\section*{Ethical Statement}

All images used in this study were anonymized before release in compliance with data privacy and ethical guidelines. Since the Kvasir dataset is publicly available and fully de-identified, no additional institutional ethical approval was required. This research was conducted in accordance with established standards for ethical medical imaging research and data protection.

\section*{Conflict of Interest}

The authors declare no conflicts of interest regarding the publication of this study.

\section*{Funding}

This work did not receive any external funding. The study was conducted using institutional resources and the computational facilities available to the authors.

\section*{Patient Consent Statement}

Patient consent was not required because the study used a publicly available, de-identified dataset containing anonymized endoscopic images.

\section*{Permission to Reproduce Material from Other Sources}

Not applicable. No figures, tables, images, or other copyrighted materials from external sources were reproduced in this manuscript.

\section*{Clinical Trial Registration}

Not applicable. This study does not involve a clinical trial.

\end{document}